\documentclass[sigconf]{acmart}
\AtBeginDocument{%
  }

\usepackage{comment}
\usepackage{multirow}
\usepackage{subfigure}
\usepackage{bm}
\usepackage[ruled,vlined]{algorithm2e}

\theoremstyle{plain}
\newtheorem{theorem}{Theorem}[section]

\theoremstyle{definition}
\newtheorem{definition}[theorem]{Definition}

\newtheorem{remark}[theorem]{Remark}

\copyrightyear{2025}
\acmYear{2025}
\setcopyright{acmlicensed}\acmConference[KDD '25]{Proceedings of the 31st ACM SIGKDD Conference on Knowledge Discovery and Data Mining V.2}{August 3--7, 2025}{Toronto, ON, Canada}
\acmBooktitle{Proceedings of the 31st ACM SIGKDD Conference on Knowledge Discovery and Data Mining V.2 (KDD '25), August 3--7, 2025, Toronto, ON, Canada}
\acmDOI{10.1145/3711896.3736924}
\acmISBN{979-8-4007-1454-2/2025/08}

\begin{document}

\title{Efficient Identity and Position Graph Embedding via Spectral-Based Random Feature Aggregation}

\author{Meng Qin}
\affiliation{%
  \institution{Department of Strategic \& Advanced Interdisciplinary Research, Pengcheng Laboratory}
  \city{Shenzhen}
  \country{China}
}
\email{mengqin\_az@foxmail.com}

\author{Jiahong Liu}
\affiliation{%
  \institution{The Chinese University of Hong Kong}
  \city{Hong Kong SAR}
  \country{China}}
\email{jiahong.liu21@gmail.com}

\author{Irwin King}
\affiliation{%
  \institution{The Chinese University of Hong Kong}
  \city{Hong Kong SAR}
  \country{China}
}
\email{king@cse.cuhk.edu.hk}

\renewcommand{\shortauthors}{Meng Qin, Jiahong Liu, \& Irwin King}

\begin{abstract}
  Graph neural networks (GNNs), which capture graph structures via a feature aggregation mechanism following the graph embedding framework, have demonstrated a powerful ability to support various tasks. According to the topology properties (e.g., structural roles or community memberships of nodes) to be preserved, graph embedding can be categorized into identity and position embedding. However, it is unclear for most GNN-based methods which property they can capture. Some of them may also suffer from low efficiency and scalability caused by several time- and space-consuming procedures (e.g., feature extraction and training). From a perspective of graph signal processing, we find that high- and low-frequency information in the graph spectral domain may characterize node identities and positions, respectively. Based on this investigation, we propose random feature aggregation (RFA) for efficient identity and position embedding, serving as an extreme ablation study regarding GNN feature aggregation. RFA (\romannumeral1) adopts a spectral-based GNN without learnable parameters as its backbone, (\romannumeral2) only uses random noises as inputs, and (\romannumeral3) derives embeddings via just one feed-forward propagation (FFP). Inspired by degree-corrected spectral clustering, we further introduce a degree correction mechanism to the GNN backbone. Surprisingly, our experiments demonstrate that two variants of RFA with high- and low-pass filters can respectively derive informative identity and position embeddings via just one FFP (i.e., without any training). As a result, RFA can achieve a better trade-off between quality and efficiency for both identity and position embedding over various baselines. We have made our code public at https://github.com/KuroginQin/RFA
\end{abstract}





\maketitle

\section{Introduction}\label{Sec:Intro}
Graph neural networks (GNNs) have emerged as a state-of-the-art technique to support various graph inference tasks 
(e.g., node classification \cite{kipf2022semi,velivckovicdeep}, community detection \cite{qin2024pre,wu2024procom,qin2024towards}, and link prediction \cite{zhang2018link,qin2023temporal,qin2023high}).
Following the \textit{\textbf{graph embedding}} framework, which maps nodes to low-dimensional vector representations (a.k.a. embeddings) with specific graph properties preserved, most GNNs capture non-Euclidean graph structures and derive corresponding embeddings via the feature aggregation (a.k.a. message passing) mechanism.
According to the topology properties to be preserved, graph embedding can be categorized into \textit{\textbf{identity}} and \textit{\textbf{position embedding}} \cite{rossi2020proximity,zhu2021node,qin2024irwe}.

\begin{figure}[t]
  \centering
  \includegraphics[width=\linewidth, trim=22 30 20 20,clip]{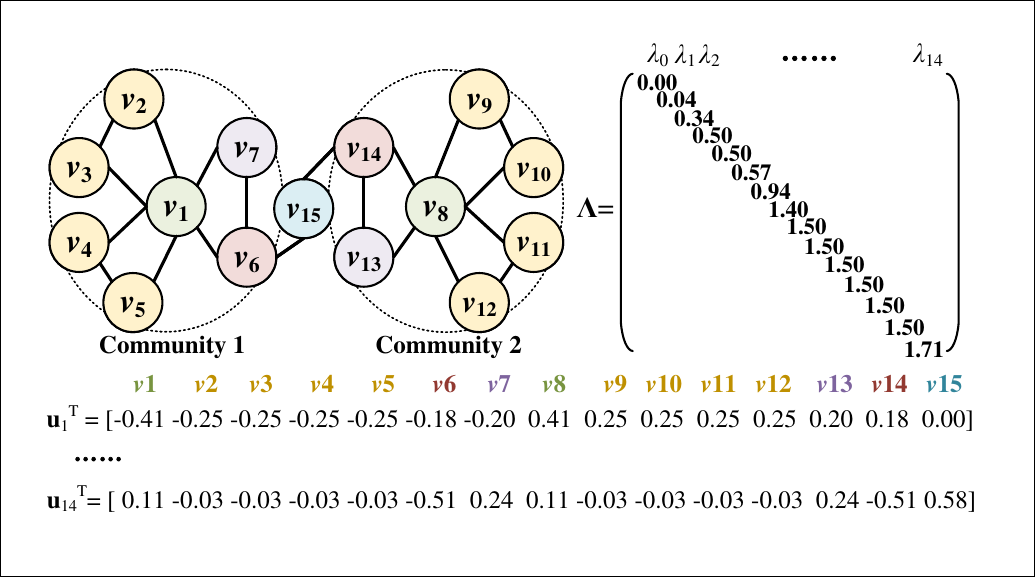}
  \vspace{-0.6cm}
  \caption{An example of node identities and positions as well as ED of graph Laplacian (see Fig.~\ref{Fig:Toy-1-vecs} for the full eigenvectors), where each color denotes a unique identity; nodes in the same community have similar positions.
  }\label{Fig:Toy}
  \vspace{-0.4cm}
\end{figure}

The identity embedding (a.k.a. structural embedding) captures the role (or identity) of each node in graph topology \cite{ribeiro2017struc2vec}, characterized by its rooted subgraph. For instance, nodes with the same color in Fig.~\ref{Fig:Toy} are expected to play the same role, where green (e.g., $\{v_1, v_8\}$) and yellow (e.g., $\{v_2, v_9\}$) may indicate the opinion leader and ordinary audience in a social network \cite{yang2015rain}. Typical identity embedding approaches include \textit{struc2vec} \cite{ribeiro2017struc2vec} and \textit{GraphWave} \cite{donnat2018learning}.

The position embedding (a.k.a. proximity-preserving embedding) captures node positions measured by overlaps of local neighbors that characterize the community structure \cite{fortunato202220}. For instance, there are two densely connected clusters (i.e., communities) in Fig.~\ref{Fig:Toy}. Nodes in the same community (e.g., $\{v_1, \cdots, v_7\}$) have similar positions, corresponding to their high overlaps of neighbors. \textit{DeepWalk} \cite{perozzi2014deepwalk} and \textit{node2vec} \cite{grover2016node2vec} are classic position embedding methods.

Note that node identities and positions are different types of properties, which may contradict with each other. For instance, two nodes can have the same identity even though they are in different communities with a low overlap of neighbors (e.g., $\{v_1, v_8\}$ in Fig.~\ref{Fig:Toy}).

While GNN-based approaches \cite{velivckovicdeep,hou2022graphmae,hou2023graphmae2} have demonstrated a powerful ability to derive informative embeddings for several inference tasks, most of them are originally designed for attributed graphs \cite{liu2022discovering,liu2022cspm}.
As validated by \cite{newman2016structure,qin2018adaptive,wang2020gcn,qin2021dual}, there exist complicated correlations between graph topology and attributes, where attributes may provide (\romannumeral1) \textit{additional information beyond topology for better quality of downstream tasks} and also (\romannumeral2) \textit{inconsistent characteristics causing unexpected quality degradation}.
Experiments of \cite{qin2024irwe} illustrated that (\romannumeral1) \textit{attributes may fail to capture neither node positions nor identities}, which are pure topology properties; (\romannumeral2) \textit{the integration of attributes may even damage the quality of position or identity embeddings}.
Hence, it is unclear for some GNN-based methods which topology properties (i.e., node identities or positions) they can capture. In our experiments, these approaches may even fail to outperform conventional topology-based embedding baselines (e.g., \textit{node2vec} and \textit{struc2vec}) for identity and position embedding.

Recent studies \cite{you2019position,jin2020gralsp,you2021identity,chen2023csgcl} have tried to enable GNNs to capture node identities or positions by revising their attribute inputs or feature aggregation mechanisms. However, the effectiveness of these techniques relies heavily on additional steps about graph topology, e.g., random walk (RW) sampling and anonymous walk (AW) mapping in \textit{GraLSP} \cite{jin2020gralsp}; anchor node sampling and distance computation in \textit{P-GNN} \cite{you2019position}. These steps are then combined with a training procedure (e.g., based on gradient descent), which is usually time- and space-consuming.
Our experiments also demonstrate that these GNN-based approaches may easily suffer from low efficiency and scalability on large graphs.

In this study, we focus on \textit{\textbf{efficient unsupervised graph embedding}} and explore the potential of \textit{\textbf{GNN feature aggregation to capture node identities and positions}}.
Following early studies on identity and position embedding \cite{perozzi2014deepwalk,grover2016node2vec,ribeiro2017struc2vec,donnat2018learning}, we assume that topology is the only available input (i.e., without graph attributes). It eliminates the unclear influence from attributes to the two topology properties (i.e., node identities and positions).
We start our study by presenting the following investigation regarding the two properties from a perspective of graph signal processing (GSP) \cite{shuman2013emerging}.

\textbf{Investigation on Node Identities \& Positions: A GSP View}.
Let ${\bf{A}}$ be the adjacency matrix of a graph. In GSP, we first derive the normalized graph Laplacian ${\bf{L}}$ w.r.t. ${\bf{A}}$ and conduct the eigen-decomposition (ED) on ${\bf{L}}$.
The eigenvalues $\{ \lambda_r \}$ of ${\bf{L}}$, which are usually arranged in the order $0 = \lambda_0 \le \lambda_1 \le \cdots, \le \lambda_{N-1} \le 2$ as in Fig.~\ref{Fig:Toy}, are defined as \textit{\textbf{frequencies}} of a graph with $N$ nodes.
Let ${\bf{u}}_r \in \mathbb{R}^N$ be the eigenvector of $\lambda_r$. The \textit{\textbf{frequency}} $\lambda_r$ indicates the variation of values in ${\bf{u}}_r$ w.r.t. the local topology.
Namely, a lower frequency $\lambda_r$ implies that nodes are more likely to have similar values in ${\bf{u}}_r$ (i.e., a small variation) if they are densely connected (i.e., in the same community).
In Fig.~\ref{Fig:Toy}, one can extract the community structure based on signs of entries in ${\bf{u}}_1$ (i.e., eigenvector w.r.t. the smallest non-zero eigenvalue\footnote{The multiplicity of zero eigenvalue equals to the number of connected components \cite{shuman2013emerging}. Since the graph in Fig.~\ref{Fig:Toy} is connected with only one connected component, we have $0 = \lambda_0 < \lambda_1$. ${\bf{u}}_0$ is usually meaningless for community structures.}), where $({\bf{u}}_1)_i < 0$ (or $> 0$) for each node $v_i$ in community 1 (or community 2). We also have $({\bf{u}}_1)_{15} = 0$ as it is unclear for $v_{15}$ which community it belongs to.
Consistent with this observation, the well-known spectral clustering \cite{von2007tutorial} uses eigenvectors w.r.t. the smallest $K$ eigenvalues as the input of $K$Means to derive a $K$-way community partition.
Hence, \textit{eigenvectors w.r.t. small eigenvalues (i.e., low-frequency information in the graph spectral domain) may capture node positions}.

In contrast, the capacity of high-frequency information to capture topology properties is seldom explored.
Surprisingly, we find that \textit{high-frequency information may capture node identities}. In ${\bf{u}}_{14}$ w.r.t the largest eigenvalue $\lambda_{14}$ (i.e., the highest frequency) of Fig.~\ref{Fig:Toy}, nodes with the same identity have exactly the same value (e.g., $0.11$ for green nodes and $-0.03$ for yellow nodes).

In summary, \textit{two extremes (i.e., high- and low-frequency information) of the graph spectral domain may have the potential to characterize node identities and positions, respectively}.
Please refer to Appendix~\ref{App:Toy} for more examples to validate this conclusion.

\textbf{Presented Work}.
Based on the aforementioned investigation, we propose random feature aggregation (RFA), a simple yet effective method for efficient identity and position embedding.

\textit{\textbf{An ablation study about GNN feature aggregation}}.
Different from most GNN-based approaches, RFA adopts a spectral-based GNN as its backbone but removes all the learnable model parameters.
Motivated by efficient embedding techniques with random projection \cite{zhang2018billion,bhowmick2020louvainne,gao2023raftgp}, we only feed random noises to the GNN backbone and derive embeddings via just one feed-forward propagation (FFP). In this setting, RFA serves as an extreme ablation study about the feature aggregation of spectral-based GNNs.
Surprisingly, we find out that two variants of RFA, which use high- and low-pass filters, can derive informative identity and position embeddings.

\textit{\textbf{A new spectral-based feature aggregation mechanism}}.
Inspired by recent advances in degree-corrected spectral clustering (DCSC) \cite{qin2013regularized,jin2021improvements}, we introduce a degree correction mechanism to each GNN layer of RFA, which can adjust the distribution of graph frequencies $\{ \lambda_r \}$. To the best of our knowledge, we are the first to introduce degree correction to spectral-based GNNs.
The modified graph convolution in each layer is further combined with non-linear activation and normalization.

\textit{\textbf{A better trade-off between quality and efficiency}}.
In summary, RFA is efficient without any feature extraction, ED, or training procedures while having the potential to capture node identities and positions based on the modified GNN backbone.
Experiments on datasets with various scales demonstrate that RFA can achieve a better trade-off between the inference quality and efficiency for both identity and position embedding over various baselines.

\section{Problem Statements \& Preliminaries}\label{Sec:Prob}
In this study, we consider \textit{\textbf{unsupervised graph embedding}} on undirected and unweighted graphs, with topology structures as the only available input (i.e., attributes are unavailable).
In general, one can represent a graph as $G = (V, E)$, with $V := \{ v_1, v_2, \cdots, v_N\}$ and $E := \{ (v_i, v_j) | v_i, v_j \in V\}$ as the sets of nodes and edges.
The topology of $G$ can be described by an adjacency matrix ${\bf{A}} \in \{ 0, 1\} ^ {N \times N}$, where ${\bf{A}}_{ij} = {\bf{A}}_{ji} = 1$ if $(v_i, v_j) \in E$ and ${\bf{A}}_{ij} = {\bf{A}}_{ji} = 0$, otherwise.
Node identity and position are two typical types of properties regarding graph topology.
\begin{definition}[\textbf{Node Identity}]\label{Def:Ide}
    For each node $v_i \in V$, node identity describes the structural role that $v_i$ plays in graph topology, which can be characterized by the $l$-hop sub-graph $G_l(v_i)$ rooted at $v_i$ (a.k.a. the ego-net of $v_i$). Usually, nodes $(v_i, v_j)$ with similar ego-nets $( G_l (v_i), G_l (v_j) )$ should play similar structural roles and thus have similar identities.
\end{definition}
\begin{definition}[\textbf{Node Position}]\label{Def:Pos}
    In a graph $G$, positions of nodes $\{ v_i \}$ can be encoded the overlaps of their $l$-hop neighbors (i.e., the community structure \cite{fortunato202220}). Concretely, nodes $\{ v_i \}$ with high overlaps of $l$-hop neighbors are expected to belong to the same community and thus have similar positions.
\end{definition}
In Fig.~\ref{Fig:Toy}, $v_1$ ad $v_8$ are expected to play the same structural role (e.g., opinion leader in a social network \cite{yang2015rain}) since their $1$-hop ego-nets $G_1 (v_1)$ and $G_1 (v_1)$ induced by $\{ v_1 \cdots, v_7\}$ and $\{ v_8, \cdots, v_{14}\}$ have the same topology. Namely, $v_1$ and $v_8$ should have the same identity. Moreover, $\{ v_1, \cdots, v_7 \}$ belong to the same community (i.e, community $1$) with high overlaps of $1$-hop neighbors and thus should have similar positions.

\begin{definition}[\textbf{Graph Embedding}]\label{Def:Emb}
    Given a graph $G$, graph embedding can be formulated as a function $f:V \to {\mathbb{R}^d}$ that maps each node $v_i \in V$ to a low-dimensional vector (i.e., embedding) ${\bf{z}}_i \in \mathbb{R}^d$ ($d \ll N$) with specific graph properties preserved. The derived embeddings $\{ {\bf{z}}_i \}$ are usually adopted as inputs of some downstream modules to support concrete inference tasks.
    %(e.g., logistic regression for node classification).
    We define ${\bf{z}}_i := f (v_i)$ as the \textit{\textbf{identity}} (or \textit{\textbf{position}}) \textit{\textbf{embedding}}, if $f$ preserves node identities (or positions), where nodes $(v_i, v_j)$ with similar identities (or positions) should have close embeddings $({\bf{z}}_i, {\bf{z}}_j)$ in terms of a short distance or high similarity.
\end{definition}

Given ${\bf{A}}$, we can get the normalized graph Laplacian ${\bf{L}} := {\bf{I}}_N - {\bf{D}}^{-1/2}{\bf{A}}{\bf{D}}^{-1/2}$, where ${\bf{I}}_N$ denotes an $N$-dimensional identity matrix; ${\bf{D}} := {\mathop{\rm diag}\nolimits} ({\deg}_1, \cdots, {\deg}_{N})$ is the degree diagonal matrix w.r.t. ${\bf{A}}$; ${\deg}_i = \sum\nolimits_j {{{\bf{A}}_{ij}}} $ is the degree of node $v_i$.
Since ${\bf{L}}$ is symmetric and positive semi-definite \cite{von2007tutorial}, it has $N$ orthonormal eigenvectors $\{ {\bf{u}}_i \}_{i=0}^{N-1} \in \mathbb{R}^N$, which can be arranged as ${\bf{U}}: = [{{\bf{u}}_0},{{\bf{u}}_1}, \cdots {{\bf{u}}_{N - 1}}]$. Let $\lambda_i$ be the eigenvalue of ${\bf{u}}_i$, we arrange all the eigenvalues as a diagonal matrix ${\bf{\Lambda }}: = {\mathop{\rm diag}\nolimits} ({\lambda _0},{\lambda _1}, \cdots ,{\lambda _{N - 1}})$, with $0 = {\lambda _0} \le {\lambda _1} \le  \cdots  \le {\lambda _{N - 1}} \le 2$.
ED on ${\bf{L}}$ can be represented as ${\bf{L}} = {\bf{U\Lambda }}{{\bf{U}}^T}$.
In GSP, $\{ \lambda_r \}$ with this setting are defined as \textit{\textbf{frequencies}} indicating the variation of values in $\{ {\bf{u}}_r \}$ w.r.t. graph topology, as shown in Fig~\ref{Fig:Toy}.
Assume that $G$ is connected with only one connected component. We have $0 = \lambda_0 < \lambda_1 \le \lambda_2 \le \cdots$.

\begin{definition}[\textbf{Graph Convolution}]\label{Def:GF}
    Given a signal ${\bf{z}} \in \mathbb{R}^{N}$ to be processed, the \textit{\textbf{graph Fourier transform}} is defined as ${\bf{\hat z}} =  \mathcal{F} ({\bf{z}}) := {\bf{U}}^T {\bf{z}}$, which treats orthogonal $\{ {\bf{u}}_i \}$ as bases and maps ${\bf{z}}$ from the spatial domain to spectral domain. The \textit{\textbf{inverse graph Fourier transform}} is defined as ${\bf{z}} = \mathcal{F}^{-1} ({\bf{\hat z}}) := {\bf{U}} {\bf{\hat z}}$, which recovers ${\bf{\hat z}}$ from the spectral domain to spatial domain. Let $g({\bf{\Lambda}})$ be the \textit{\textbf{graph convolution kernel}}, a function regarding ${\bf{\Lambda}}$. The \textit{\textbf{graph convolution}} operation can be described as $g * {\bf{z}} := {\bf{U}} g({\bf{\Lambda}}) {\bf{U}}^T \bf{z}$.
\end{definition}

\section{Methodology}\label{Sec:Meth}
Based on our investigation in Fig.~\ref{Fig:Toy} and Appendix~\ref{App:Toy}, we propose RFA, which serves as an extreme ablation regarding the potential of GNN feature aggregation to capture topology properties.
Concretely, we use a spectral-based GNN without learnable parameters as the backbone. Inspired by the efficient random projection technique \cite{arriaga2006algorithmic}, we only feed random noises to this backbone and derive embeddings via just one FFP, without any additional feature extraction, ED, and training.

\subsection{Spectral-Based GNN Feature Aggregation}
Motivated by prior studies on spectral-based GNNs \cite{bo2021beyond,dong2021adagnn}, we consider the following graph convolution kernel
\begin{equation}\label{Eq:Filter}
    g ({\bf{\Lambda}}): = (\delta  + \alpha ){{\bf{I}}_N} - \alpha {\bf{\Lambda }},
\end{equation}
with $\delta \in [0, 1]$ and $\alpha \in [-1, 1]$ as two hyper-parameters.
When $\alpha > 0$, $g ({\bf{\Lambda}})$ corresponds to a monotone decreasing function regarding $\lambda$, which amplifies low-frequency signals (i.e., larger weights for bases $\{ {\bf{u}}_r \}$ w.r.t. low frequencies), and thus represents a \textit{\textbf{low-pass filter}}. In contrast, $g ({\bf{\Lambda}})$ becomes a \textit{\textbf{high-pass filter}} amplifying high-frequency signals, when $\alpha < 0$.
Namely, \textit{the sign of $\alpha$ determines that $g (\bf{\Lambda})$ is a \textbf{low-pass} or \textbf{high-pass filter}}.

Based on (\ref{Eq:Filter}) and the orthogonal property of $\{ {\bf{u}}_r \}$ (i.e., ${\bf{U}}{\bf{U}}^T = {\bf{U}}^T{\bf{U}} = {\bf{I}}_N$), the graph convolution operation can be formulated as
\begin{equation}\label{Eq:GC}
    g * {\bf{z}} = {\bf{U}} g({\bf{\Lambda}}) {{\bf{U}}^T}{\bf{z}} = \delta {\bf{z}} + \alpha {{\bf{D}}^{ - 1/2}}{\bf{A}}{{\bf{D}}^{ - 1/2}}{\bf{z}}.
\end{equation}
\begin{remark}[]
    The application of low-pass (or high-pass) filter in the spectral domain is equivalent to the \textit{\textbf{sum}} of (or \textit{\textbf{difference}} between) the (\romannumeral1) original signal ${\bf{z}}$ and (\romannumeral2) corresponding `mean' value ${\bf{D}}^{-1/2}{\bf{A}}{\bf{D}}^{-1/2}{\bf{z}}$ w.r.t. $1$-hop neighbors in the spatial domain. $\{ \delta, \alpha \}$ adjust the relative contributions of these two parts.
\end{remark}
Here, we use two extreme settings of $(\delta, \alpha) = (0.1, 1)$ and $(0.1, -1)$ for low- and high-pass filters, forming two variants of RFA.
Consistent with our investigation in Fig.~\ref{Fig:Toy} and Appendix~\ref{App:Toy}, our experiments validate that these two variants with high- and low-pass filters can derive informative identity and position embeddings.

\subsection{Degree Correction Mechanism}
It is well-known that there exists a close relation between GSP and spectral clustering.
For instance, vanilla spectral clustering \cite{von2007tutorial} extracts eigenvectors w.r.t. the smallest eigenvalues of normalized graph Laplacian ${\bf{L}}$ (i.e., low-frequency information).
As an advanced extension of this algorithm, DCSC \cite{qin2013regularized,jin2021improvements} conducts ED on the regularized graph Laplacian ${\bf{L}}_{\tau} := {\bf{I}}_N - {\bf{D}}_{\tau}^{-1/2}{\bf{A}}{\bf{D}}_{\tau}^{-1/2}$, where ${\bf{D}}_\tau := {\bf{D}} + \tau {\bf{I}}_N$; $\tau > 0$ is defined as the degree correction term.
In the rest of this paper, we use ${\bf{L}}_{\tau} = {\bf{\tilde U}} {\bf{\tilde \Lambda}} {\bf{\tilde U}}^T$ to represent the ED on ${\bf{L}}_{\tau}$, with the corresponding eigenvalues and eigenvectors arranged as ${\bf{\tilde \Lambda}} := {\mathop{\rm diag}\nolimits} ({\tilde \lambda}_0, {\tilde \lambda}_1, \cdots, {\tilde \lambda}_{N-1})$ s.t. $\tilde \lambda_0 \le \tilde \lambda_1 \le \cdots \le \tilde \lambda_{N-1}$ and ${\bf{\tilde U}} := [{\bf{\tilde u}}_0, {\bf{\tilde u}}_1, \cdots, {\bf{\tilde u}}_{N-1}]$.
Note that we still have ${\bf{\tilde U}} {\bf{\tilde U}}^T = {\bf{\tilde U}}^T {\bf{\tilde U}} = {\bf{I}}_N$.

\begin{figure}[]
 \begin{minipage}{0.31\linewidth}
 \subfigure[$\tau = 0$]{
  \includegraphics[width=\textwidth,trim=8 8 25 15,clip]{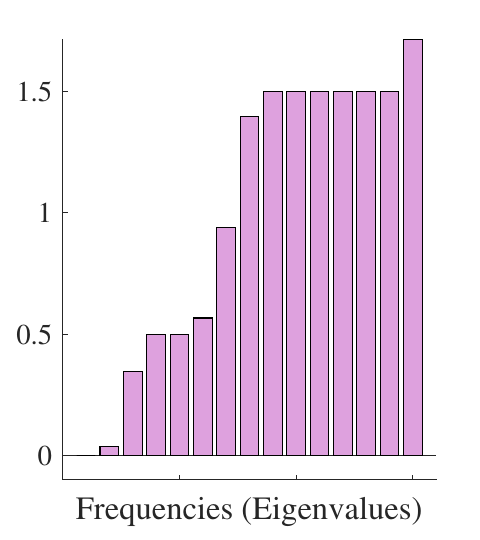}
  }
 \end{minipage}
 \begin{minipage}{0.31\linewidth}
 \subfigure[$\tau = 1$]{
  \includegraphics[width=\textwidth,trim=8 8 25 15,clip]{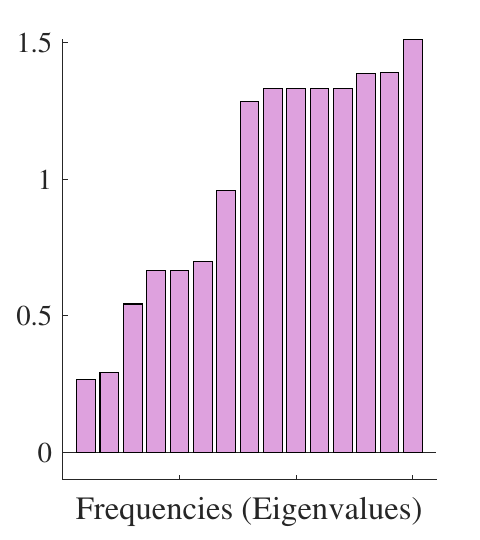}
  }
 \end{minipage}
 \begin{minipage}{0.31\linewidth}
 \subfigure[$\tau = 5$]{
  \includegraphics[width=\textwidth,trim=8 8 25 15,clip]{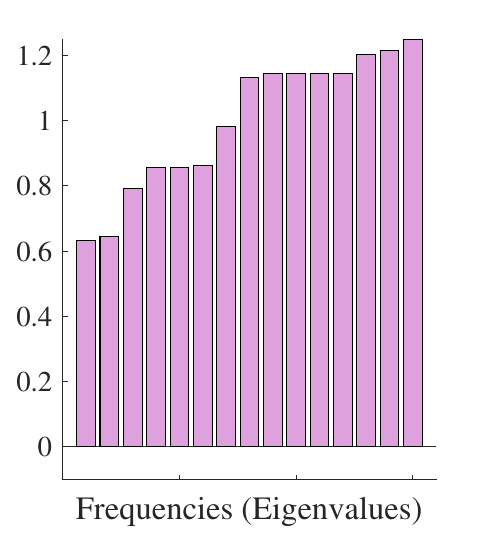}
  }
 \end{minipage}
 \begin{minipage}{0.31\linewidth}
 \subfigure[$\tau = 10$]{
  \includegraphics[width=\textwidth,trim=8 8 25 15,clip]{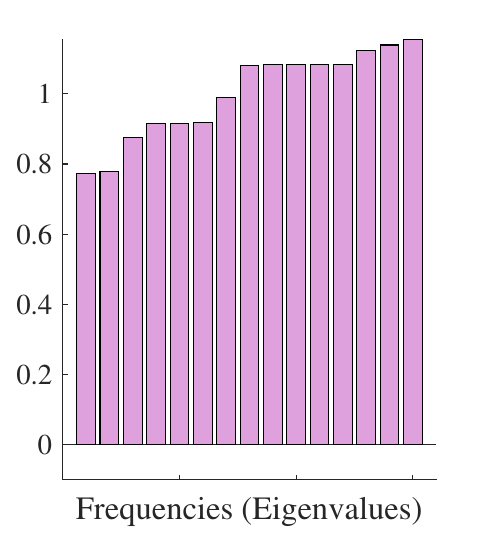}
  }
 \end{minipage}
 \begin{minipage}{0.31\linewidth}
 \subfigure[$\tau = 50$]{
  \includegraphics[width=\textwidth,trim=8 8 25 15,clip]{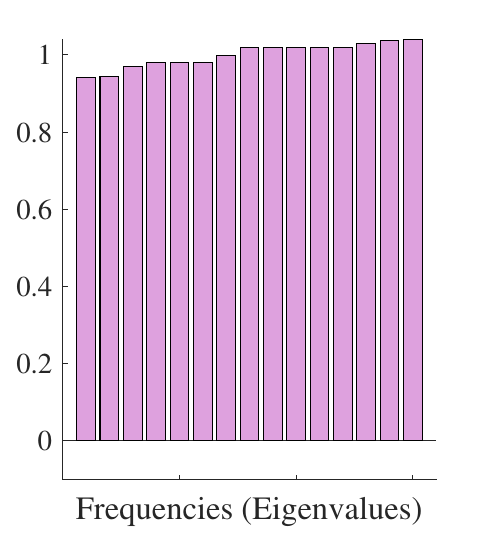}
  }
 \end{minipage}
 \begin{minipage}{0.31\linewidth}
 \subfigure[$\tau = 100$]{
  \includegraphics[width=\textwidth,trim=8 8 25 15,clip]{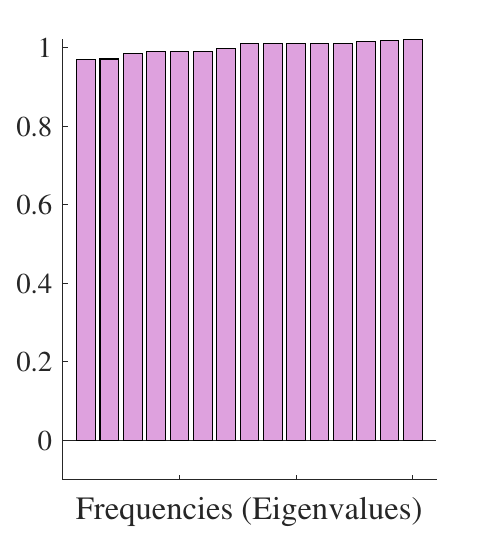}
  }
 \end{minipage}
 %----------
 \vspace{-0.4cm}
 \caption{Distributions of frequencies $\{ \tilde \lambda_r \}$ w.r.t. different settings of $\tau$ based on the graph in Fig.~\ref{Fig:Toy}.
 }\label{Fig:Case-Tau}
 \vspace{-0.4cm}
\end{figure}

Inspired by recent advances in DCSC, we extend the graph convolution operation defined in (\ref{Eq:GC}) to the following form:
\begin{equation}\label{Eq:RGC}
    g * {\bf{z}} = {\bf{\tilde U}} g({\bf{\tilde \Lambda}}) {{\bf{\tilde U}}^T}{\bf{z}} = \delta {\bf{z}} + \alpha {{\bf{D}}_{\tau}^{ - 1/2}}{\bf{A}}{{\bf{D}}_{\tau}^{ - 1/2}}{\bf{z}},
\end{equation}
where we treat $\tau$ as a tunable hyper-parameter. It is equivalent to applying $g({\bf{\tilde \Lambda }})$ in the spectral domain.
The following \textit{\textbf{Gershgorin Circle theorem}} helps interpret the effect of introducing $\tau$.
\begin{theorem}[\textbf{Gershgorin Circle Theorem} \cite{barany2017gershgorin}]\label{Th:Disk}
    The eigenvalues of a matrix ${\bf{M}} \in \mathbb{C}^{N \times N}$ lie in the union of $N$ discs $(\mathcal{D}_1, \cdots, \mathcal{D}_N)$, where $\mathcal{D}_i := x \in \mathbb{C} : |x - {\bf{M}}_{ii}| \le r_i$; $r_i := \sum\nolimits_{j = 1,j \ne i}^N {|{{\bf{M}}_{ij}}|}$.
\end{theorem}
\begin{remark}[]\label{Re:tau}
    By \textbf{Theorem~\ref{Th:Disk}}, eigenvalues of ${\bf{I}}_N - {\bf{D}}_{\tau}^{-1} {\bf{A}}$ (analogous to ${\bf{L}}_\tau = {\bf{I}}_N - {\bf{D}}_{\tau}^{-1/2} {\bf{A}} {\bf{D}}_{\tau}^{-1/2}$) lie in circles centered at $1$, whose radiuses are controlled by $\{ {\deg}_i / ({\deg}_i + \tau) \}$. In this sense, $\tau$ can adjust the distribution of eigenvalues $\{ \tilde \lambda_r \}$ (i.e., frequencies in the graph spectral domain) and further affect eigenvectors $\{ {\bf{\tilde u}}_r \}$.
\end{remark}
Based on the example graph in Fig.~\ref{Fig:Toy}, Fig.~\ref{Fig:Case-Tau} visualizes corresponding eigenvalues $\{ \tilde \lambda_r \}$ w.r.t. $\tau \in \{0, 1, 5, 10, 50, 100\}$. In Fig.~\ref{Fig:Case-Tau}, the increase of $\tau$ will force $\{ \tilde \lambda_r\}$ to have close values. For a large $\tau$ (e.g., $\tau >> 100$), all the eigenvalues $\{ \tilde \lambda_r\}$ approach to $1$. This observation is consistent with \textbf{Remark~\ref{Re:tau}}, where $\{ \tilde \lambda_r\}$ lie in circles centered at $1$ and with radiuses of $\mathop {\lim }\limits_{\tau  \to \infty } {\deg _i}/(\tau  + {\deg _i}) = 0$.

Surprisingly, our experiments demonstrate that a relatively small setting of $\tau$ (e.g., $\tau = 20$) can significantly improve the embedding quality of RFA compared with $\tau = 0$ (i.e., directly using the classic convolution operation defined in (\ref{Eq:GC})).
To the best of our knowledge, we are the first to introduce the aforementioned degree correction mechanism to spectral-based GNNs.

\begin{algorithm}[t]\small
\caption{\small Inference Algorithm of RFA}
\label{Alg:RFA}
\LinesNumbered
\KwIn{input graph $G = (V, E)$; hyper-parameters $\{ \tau, K\}$; embedding dimensionality $d$}
\KwOut{derived embeddings ${\bf{Z}}$}
generate random noise input via $\bm{\Theta} \in \mathbb{R}^{N \times d} \sim \mathcal{N} (0, 1/d)$\\
${{\bf{Z}}^{(0)}} \leftarrow {\bm{\Theta}} $ \\
\For{$k$ \bf{from} $1$ \bf{to} $K$}
{
    ${{\bf{Z}}^{(k)}} \leftarrow {\rm{Norm}}({f_{{\rm{act}}}}([\delta {{\bf{I}}_N} + \alpha {\bf{D}}_\tau ^{ - 1/2}{\bf{AD}}_\tau ^{ - 1/2}]{{\bf{Z}}^{(k - 1)}}))$\\
}
${\bf{Z}} \leftarrow {\bf{Z}}^{(K)}$ \\
\end{algorithm}

\subsection{Model Architecture \& Inference Algorithm}
To build a feasible spectral-based GNN backbone for RFA, we can adopt (\ref{Eq:RGC}) as the operation of one layer and stack multiple such layers.
However, one FFP on this architecture is not enough for RFA to derive informative embeddings and outperform most baselines, as illustrated in our ablation study.
Surprisingly, we found that the integration of (\romannumeral1) non-linear activation and (\romannumeral2) normalization, which are widely used in existing GNN-based approaches, can help RFA achieve impressive quality for identity and position embedding.

Let ${\bf{Z}}^{(k-1)} \in \mathbb{R}^{N \times d}$ and ${\bf{Z}}^{(k)} \in \mathbb{R}^{N \times d}$ be the input and output of the $k$-th layer in RFA. Since we only feed random noises to RFA, we set ${\bf{Z}}^{(0)} = {\bm{\Theta}} \in \mathbb{R}^{N \times d} \sim \mathcal{N} (0, 1/d)$. Following the modified graph convolution operation described in (\ref{Eq:RGC}), the $k$-th layer is defined as
\begin{equation}\label{Eq:RFA}
    {{\bf{Z}}^{(k)}} = {\mathop{\rm Norm}\nolimits} ( {f_{{\rm{act}}}}( [\delta {{\bf{I}}_N} + \alpha {\bf{D}}_\tau ^{ - 1/2}{\bf{AD}}_\tau ^{ - 1/2}]{{\bf{Z}}^{(k - 1)}} )),
\end{equation}
where $ {f_{{\rm{act}}}} (\cdot)$ denotes a non-linear activation function; ${\mathop{\rm Norm}\nolimits} (\cdot)$ is an optional normalization operation.
Suppose that there are $K$ GNN layers. We use ${\bf{Z}} = {\bf{Z}}^{(K)} \in \mathbb{R}^{N \times d}$ to represent the final embeddings given by RFA, with ${\bf{Z}}_{i,:} = {\bf{z}}_i \in \mathbb{R}^d$ as the embedding of node $v_i$.

In this study, $\tanh (\cdot)$ and $\exp (\cdot)$ are adopted as two examples of $ {f_{{\rm{act}}}} (\cdot)$. Moreover, we also try two settings for ${\mathop{\rm Norm}\nolimits} (\cdot)$. Let ${\bf{M}}$ be a matrix to be normalized. We first consider the (row-wise) $l2$-normalization that updates each row of ${\bf{M}}$ via ${{\bf{M}}_{i,:}} \leftarrow {{\bf{M}}_{i,:}}/|{{\bf{M}}_{i,:}}|{_2}$. We also consider the z-score normalization. It updates each column of ${\bf{M}}$ via ${{\bf{M}}_{:,j}} \leftarrow ({{\bf{M}}_{:,j}} - \mu ({{\bf{M}}_{:,j}}))/\sigma ({{\bf{M}}_{:,j}})$, where $\mu ({{\bf{M}}_{:,j}}))$ and $\sigma ({{\bf{M}}_{:,j}})$ denote the mean and standard deviation of ${\bf{M}}_{:,j}$.

A reasonable interpretation of why the non-linear activation and normalization can ensure the embedding quality of RFA is that \textit{they may control the magnitude of feature aggregation outputs and make the derived embeddings more distinguishable}. Instead, directly stacking (\ref{Eq:RGC}) may force the magnitude of some outputs to be (\romannumeral1) extremely large causing the numerical overflow exception or (\romannumeral2) extremely small and not distinguishable for downstream modules. Concretely, the output range of $\tanh (\cdot)$ is $[-1, 1]$, which can prevent the magnitude of embeddings from being extremely large. $\exp (\cdot)$ may help amplify some embedding values and prevent the magnitude from being extremely small.

Let $d$ be the embedding dimensionality.
Algorithm~\ref{Alg:RFA} summarizes the overall inference procedure of RFA, where we only feed random noises to the GNN backbone (i.e., lines 1-2) and derive feasible embeddings ${\bf{Z}}$ via just one FFP (i.e., lines 3-5).
In this sense, each GNN layer serves as one iteration to update embeddings.
Let $N$ and $M$ be the numbers of nodes and edges. We usually have $K < d \ll N < M$ for large sparse graphs.
By using the efficient sparse-dense matrix multiplication, the complexity of one iteration is $O ((M + N)d)$.
Then, the overall time complexity of RFA is no more than $O((Md + Nd)K) \approx O(N + M)$.
Since there are no additional learnable parameters in RFA, the space complexity is $O(Nd)$.

\subsection{Theoretical Interpretations}
In the rest of this section, we try to provide possible theoretical interpretations about the potential of low- and high-frequency information to capture node positions and identities, which correspond to RFA variants with low- and high-pass filters, respectively.

\begin{remark}[]\label{Re:Low}
    Low-frequency information in the graph spectral domain encodes key information about community structures and thus can characterize node positions.
\end{remark}

\textbf{Proof of Remark~\ref{Re:Low}}. Normalized cut (NCut) \cite{von2007tutorial} is a well-known objective for community detection. For a $C$-way community partition $(R_1, R_2, \cdots, R_C)$ (with $R_r \in V$ as the $r$-th community), it can be formulated as the following matrix form:
\begin{equation}
    {\min _{\bf{Y}}}~{\mathop{\rm tr}\nolimits} ({{\bf{Y}}^T}{\bf{LY}}) {\rm{~s.t.~}} {{\bf{Y}}_{ir}} = \left\{ {\begin{array}{*{20}{l}}
    {{{{\mathop{\rm vol}\nolimits} }^{-.5}}({R_r}),~{v_i} \in {R_r}}\\
    {0,{\rm{~otherwise}}}
    \end{array}} \right.,
\end{equation}
where ${\bf{Y}} \in \mathbb{R}^{N \times K}$ is the community membership indicator with discrete constraints; ${\mathop{\rm vol}\nolimits} ({R_r}): = \sum\nolimits_{{v_i} \in {R_r},{v_j} \in V} {{{\bf{A}}_{ij}}} $. By relaxing ${\bf{Y}}$ to be continuous values, we can derive the following relaxed objective:
\begin{equation}
    {\min}_{\bf{Y}}~{\mathop{\rm tr}\nolimits} ({{\bf{Y}}^T}{\bf{LY}}) {\rm{~s.t.~}} {{\bf{Y}}^T}{\bf{Y}} = {{\bf{I}}_K}.
\end{equation}
It can be solved via the objective with a Lagrange multiplier ${\bf{\Phi}}$:
\begin{equation}\label{Eq:Rel-NCut}
    \min \mathcal{J}: = {\mathop{\rm tr}\nolimits} ({{\bf{Y}}^T}{\bf{LY}}) - {\rm{tr}}({\bf{\Phi }}({{\bf{Y}}^T}{\bf{Y}} - {{\bf{I}}_K})).
\end{equation}
By letting $\partial \mathcal{J}/\partial {\bf{Y}} = 0$, we have ${\bf{LY}} = {\bf{Y\Phi }}$. It indicates that \textit{when ${\bf{\Phi}} = {\mathop{\rm diag}\nolimits} ({\lambda _0},{\lambda _1}, \cdots ,{\lambda _{K - 1}})$ w.r.t. ${\bf{L}}$'s smallest $K$ eigenvalues and ${\bf{Y}} = [{\bf{u}}_0, {\bf{u}}_1, \cdots, {\bf{u}}_{K-1}]$, one can achieve the optimal solution to (\ref{Eq:Rel-NCut})}. In this sense, \textit{low-frequency information encodes by $({\lambda _0},{\lambda _1}, \cdots ,{\lambda _{K - 1}})$ and $[{\bf{u}}_0, {\bf{u}}_1, \cdots, {\bf{u}}_{K-1}]$}, which corresponds to the optimal solution to the relaxed NCut objective (\ref{Eq:Rel-NCut}), \textit{can capture key information about community structures and thus characterize node positions}.

\begin{remark}[]\label{Re:High}
   High-frequency information in the graph spectral domain captures more information about the variation of node degrees w.r.t. local topology and characterizes node identities based on the close relation between node identities and high-order degrees.
\end{remark}

\textbf{Proof of Remark~\ref{Re:High}}. For ${\bf{L}}: = {{\bf{I}}_N} - {{\bf{D}}^{ - 1/2}}{\bf{A}}{{\bf{D}}^{ - 1/2}}$, we have ${\bf{L}}{{\bf{u}}_r} = {\lambda _r}{{\bf{u}}_r}$. By applying this property for each node $v_i$, we obtain
\begin{equation*}
    {({{\bf{u}}_r})_i} - \sum\nolimits_{{v_j} \in N({v_i})} { {({\deg _i}{\deg _j})^{ - 0.5}} {{({{\bf{u}}_r})}_j}}  = {\lambda _r}{({{\bf{u}}_r})_i} \Rightarrow
\end{equation*}
\begin{equation}\label{Eq:Lap-High}
   \sum\nolimits_{{v_j} \in N({v_i})} {[ \deg _i^{ - 1} - {({\deg _i}{\deg _j})^{ - .5}}{({{\bf{u}}_r})_j}/{({{\bf{u}}_r})_i} ]}  = {\lambda _r},
\end{equation}
where $N (v_i)$ denotes the neighbor set of $v_i$.
In (\ref{Eq:Lap-High}), if a node $v_i$ has low degree ${\deg}_i = |N(v_i)|$, \textit{the high frequency $\lambda_r$ will force the sum of node-pair difference $|\deg _i^{ - 1} - {({\deg _i}{\deg _j})^{ - 0.5}}{({{\bf{u}}_r})_j}/{({{\bf{u}}_r})_i}|$ to be large}.
In this sense, \textit{high-frequency bases $\{ \bf{u}_r \}$ may encode more information about the variation of node degrees w.r.t. local topology}, compared with the low-frequency information.

By Definition~\ref{Def:Ide}, the identity of each node $v_i$ can be characterized by the ego-net center at $v_i$. It is well known that \textit{the Weisfeiler-Lehman (WL) graph isomorphism test \cite{leman1968reduction} has the potential to distinguish between ego-nets and thus node identities}, despite some failure cases. A typical WL test algorithm update node color label in each iteration based on degrees w.r.t. $1$-hop neighbors.
For instance, after $K$ iterations of WL test, nodes with different WL color labels may have different degree distributions w.r.t. their $K$-hop neighbors and thus different ego-nets.
By the analysis in \cite{wang2022powerful}, \textit{if a high-frequency base ${\bf{u}}_r$ is powerful enough to encode degree information with an injective mapping between 1-hop degrees and entries in ${\bf{u}}_r$} (e.g., $({\bf{u}}_{14})_1 = ({\bf{u}}_{14})_8$ in Fig.~\ref{Fig:Toy}), \textit{RFA with multiple high-pass GNN layers may play a role similar to the WL test in distinguishing between ego-nets of nodes}.

\section{Experiments}\label{Sec:Exp}

\subsection{Experiment Setup}

\begin{table}[]\footnotesize
\caption{Statistics of datasets.}\label{Tab:Data}
\vspace{-0.2cm}
\begin{tabular}{l|l|l|l|c}
\hline
\textbf{Datasets} & $N$ & $M$ & $C$ & \multicolumn{1}{l}{\textbf{Ground-truth}} \\ \hline
Europe \cite{ribeiro2017struc2vec} & 399 & 5,993 & 4 & \multirow{5}{*}{\textbf{Node Identities}} \\
USA \cite{ribeiro2017struc2vec} & 1,186 & 13,597 & 4 &  \\
Reality-Call \cite{jiao2021survey} & 6,809 & 7,680 & 3 &  \\
Actor \cite{jiao2021survey} & 7,592 & 26,553 & 4 &  \\
Film \cite{jiao2021survey} & 26,851 & 122,234 & 4 &  \\ \hline
PPI \cite{grover2016node2vec} & 3,852 & 37,841 & 50 & \multirow{5}{*}{\textbf{Node Positions}} \\
BlogCatalog \cite{grover2016node2vec} & 10,312 & 333,983 & 39 &  \\
Flickr \cite{perozzi2014deepwalk} & 80,513 & 5,899,882 & 195 &  \\
Youtube \cite{yang2012defining} & 1,134,890 & 2,987,624 & 47 &  \\
Orkut \cite{yang2012defining} & 3,072,441 & 117,185,083 & 100 &  \\ \hline
\end{tabular}
\vspace{-0.2cm}
\end{table}

\begin{table}[]\footnotesize
\caption{Details of methods to be evaluated.}\label{Tab:Meth}
\vspace{-0.2cm}
\begin{tabular}{l|ll|l|l|l}
\hline
\textbf{Methods} & \textbf{Ide} & \textbf{Pos} & \textbf{Eff} & \textbf{GNN} & \textbf{Venues} \\ \hline
node2vec \cite{grover2016node2vec} &  & $\surd$ &  &  & KDD 2016 \\
PhUSION (PhN) -PPMI \cite{zhu2021node} &  & $\surd$ &  &  & SDM 2021 \\
PaCEr(P) \cite{yan2024pacer} &  & $\surd$ &  &  & WWW 2024 \\ \hline
struc2vec \cite{ribeiro2017struc2vec} & $\surd$ &  &  &  & KDD 2017 \\
PhUSION (PhN) -HK \cite{zhu2021node} & $\surd$ &  &  &  & SDM 2021 \\
PaCEr(I) \cite{yan2024pacer} & $\surd$ &  &  &  & WWW 2024 \\ \hline
RandNE \cite{zhang2018billion} &  &  & $\surd$ &  & ICDM 2018 \\
LouvainNE (LvnNE) \cite{bhowmick2020louvainne} &  &  & $\surd$ &  & WSDM 2020 \\
ProNE \cite{zhang2019prone} &  &  & $\surd$ &  & IJCAI 2019 \\
SketchNE (SktNE) \cite{xie2023sketchne} &  &  & $\surd$ &  & TKDE 2023 \\
SketchBANE (SktBANE) \cite{wu2024time} &  &  & $\surd$ &  & TKDE 2024 \\ \hline
P-GNN \cite{you2019position} &  & $\surd$ &  & $\surd$ & ICML 2019 \\
GraLSP \cite{jin2020gralsp} & $\surd$ &  &  & $\surd$ & AAAI 2020 \\
DGI \cite{velivckovicdeep} &  &  &  & $\surd$ & ICLR 2019 \\
GraphMAE (GMAE) \cite{hou2022graphmae} &  &  & $\surd$ & $\surd$ & KDD 2022 \\
GraphMAE2 (GMAE2) \cite{hou2023graphmae2} &  &  & $\surd$ & $\surd$ & WWW 2023 \\
GGD \cite{zheng2022rethinking} &  &  & $\surd$ & $\surd$ & NIPS 2022 \\
GraphECL (GECL) \cite{xiao2024efficient} &  &  & $\surd$ & $\surd$ & ICML 2024 \\ \hline
\textbf{RFA(L)} & & $\surd$ & $\surd$ & $\Delta$ & Ours \\
\textbf{RFA(H)} & $\surd$ & & $\surd$ & $\Delta$ & Ours \\ \hline
\end{tabular}
\vspace{-0.2cm}
\end{table}

\textbf{Datasets \& Downstream Tasks}. We evaluated RFA on $10$ datasets with statistics summarized in Table~\ref{Tab:Data}, where $N$, $M$, and $C$ are the numbers of nodes, edges, and classes.
(\romannumeral1) \textit{Europe}, (\romannumeral2) \textit{USA}, (\romannumeral3) \textit{Reality-Call}, (\romannumeral4) \textit{Actor}, and (\romannumeral5) \textit{Film} are widely-used datasets that provide ground-truth of node identities for multi-class classification.
In contrast, (\romannumeral6) \textit{PPI}, (\romannumeral7) \textit{BlogCatalog}, (\romannumeral8) \textit{Flickr}, (\romannumeral9) \textit{Youtube}, and (\romannumeral10) \textit{Orkut} are with ground-truth of node positions for multi-label classification.
We used the first $5$ datasets and multi-class node classification to evaluate identity embeddings. The last $5$ datasets and multi-label node classification were adopted to evaluate position embeddings. Logistic regression was used as the downstream classifier for the two types of classification.

We found that some of these datasets are not connected with several isolated edges. For each dataset, the largest connected component was extracted for evaluation, since we consider connected topology as stated in Section~\ref{Sec:Prob}.
Appendix~\ref{App:Exp-Set} describes further details of datasets.

\textbf{Baselines}. As summarized in Table~\ref{Tab:Meth}, we compared \textbf{RFA} with $18$ unsupervised baselines, where `Ide' and `Pos' denote identity and position methods (as claimed in their original literature); `Eff' represents efficient and scalable embedding approaches; `GNN' implies that a method is based on GNNs.
For simplicity, we denote \textbf{RFA} variants with low- and high-pass filters (i.e., $(\delta, \alpha) = (0.1, 1)$ and $(0.1, -1)$) as \textbf{RFA(L)} and \textbf{RFA(H)}, respectively.

PhUSION has several variants using different proximities for different types of embeddings. As suggested by \cite{zhu2021node}, we adopted variants with heat kernel (HK) and positive point-wise mutual information (PPMI) (denoted as PhN-HK and PhN-PPMI) for identity and position embedding.
PaCEr can derive both identity and position embeddings (denoted as PaCEr(I) and PaCEr(P)) by considering correlations between the two types of embeddings.

Similar to \textbf{RFA}, RandNE, LouvainNE, and SketchBANE are efficient randomized embedding methods with random noises as inputs. ProNE and SketchNE are scalable approaches based on efficient matrix factorization (MF).

Different from \textbf{RFA}, all the GNN-based methods are originally designed for attributed graphs and involve additional training procedures. To support the graph embedding without attributes as defined in Section~\ref{Sec:Prob}, we use the one-hot degree encoding as an auxiliary attribute input for these baselines, which is a standard strategy for GNNs when attributes are unavailable.

\textbf{Evaluation Criteria}.
We used micro- and macro-F1 scores to measure the embedding quality w.r.t. the downstream classification task.
On each dataset, $20\%$ of labeled nodes were sampled to train the classifier with the remaining nodes used for testing. We repeated this procedure $10$ times and reported the mean of each quality metric.
Moreover, the overall inference time (seconds) of a method to derive feasible embeddings was adopted as the efficiency metric, with the mean over $5$ independent runs reported. In particular, we define that a method encounters the out-of-time (OOT) exception if it cannot derive embeddings within $10^4$ seconds.
Usually, larger micro- and macro-F1 scores indicate better inference quality. Smaller inference time implies better efficiency.

All the experiments were conducted on a server with one Intel Xeon Gold 6430 CPU, 120 GB main memory, one RTX 4090 GPU (24GB memory), and Ubuntu 22.04 Linux OS.
We used the official implementations of all the baselines and tuned their parameters. On each dataset, the same embedding dimensionality $d$ was set for all the methods.
Similar to \textbf{RFA}, we also tried several normalization strategies (i.e., $l2$, z-score, and without normalization) for each baseline, with the best quality metrics reported.
Since all the GNN-based approaches in Table~\ref{Tab:Meth} were implemented by either \texttt{PyTorch} or \texttt{TensorFlow}, we ran them on the GPU.
For \textbf{RFA}, we provided both CPU and GPU implementations with the better inference time reported.
Detailed parameter settings are given in Appendix~\ref{App:Exp-Set}.

\subsection{Quantitative Evaluation}\label{Sec:Eva-Res}

\begin{table*}[]\scriptsize
\caption{Evaluation results of node identity embedding in terms of inference time (sec), micro-F1 (\%), and macro-F1 (\%).}\label{Tab:Eva-Ide}
\vspace{-0.2cm}
\begin{tabular}{l|p{0.5cm}p{0.5cm}l|p{0.5cm}p{0.5cm}l|p{0.55cm}p{0.5cm}l|p{0.55cm}p{0.5cm}l|p{0.65cm}p{0.5cm}l}
\hline
\multirow{2}{*}{\textbf{}} & \multicolumn{3}{c|}{\textbf{Europe}} & \multicolumn{3}{c|}{\textbf{USA}} & \multicolumn{3}{c|}{\textbf{Reality-Call}} & \multicolumn{3}{c|}{\textbf{Actor}} & \multicolumn{3}{c}{\textbf{Film}} \\ \cline{2-16} 
 & \tiny{\textbf{Time}$\downarrow$} & \tiny{\textbf{\text{Micro-}}$\uparrow$} & \tiny{\textbf{\text{Macro-F1}}$\uparrow$} & \tiny{\textbf{Time}$\downarrow$} & \tiny{\textbf{\text{Micro-}}$\uparrow$} & \tiny{\textbf{\text{Macro-F1}}$\uparrow$} & \tiny{\textbf{Time}$\downarrow$} & \tiny{\textbf{\text{Micro-}}$\uparrow$} & \tiny{\textbf{\text{Macro-F1}}$\uparrow$} & \tiny{\textbf{Time}$\downarrow$} & \tiny{\textbf{\text{Micro-}}$\uparrow$} & \tiny{\textbf{\text{Macro-F1}}$\uparrow$} & \multicolumn{1}{l}{\tiny{\textbf{Time}$\downarrow$}} & \textbf{\tiny{\text{Micro-}}}$\uparrow$ & \textbf{\tiny{\text{Macro-F1}}}$\uparrow$ \\ \hline
node2vec & 13.0294 & 35.50 & 33.61 & 41.4918 & 52.29 & 51.23 & 249.4141 & 84.27 & 33.82 & 299.1913 & 27.61 & 30.92 & 1313.4092 & 45.35 & 30.11 \\
PhN-PPMI & 0.7183 & 31.94 & 30.79 & 5.0119 & 50.17 & 49.18 & 107.8405 & 84.42 & 38.82 & 161.5225 & 32.50 & 27.44 & 2198.7719 & 45.35 & 32.43 \\
PaCEr(P) & 2.8398 & 46.62 & 46.60 & 13.0820 & 54.71 & 53.80 & 128.8800 & 82.94 & 44.53 & 351.1945 & 39.29 & 33.43 & 6179.1326 & 47.62 & 32.91 \\ \hline
struc2vec & 12.1520 & \underline{53.00} & \underline{52.56} & 46.6643 & 56.49 & 55.77 & 213.9466 & 85.01 & \underline{48.93} & 441.5575 & 40.12 & 34.00 & 2617.2983 & 47.62 & 32.97 \\
PhN-HK & 3.7233 & 50.75 & 48.54 & 29.0471 & \textbf{58.91} & \textbf{57.98} & 800.8893 & 85.19 & 46.77 & 1184.9653 & 40.77 & \underline{36.19} & 5996.9918 & 54.41 & \underline{37.92} \\
PaCEr(I) & 1.7805 & \underline{54.97} & \underline{54.33} & 11.9669 & 56.48 & 55.07 & 122.6423 & 85.41 & 44.55 & 343.6011 & \underline{42.18} & 33.40 & 5695.4612 & 51.83 & \underline{38.96} \\ \hline
RandNE & \underline{0.0135} & 51.78 & 50.62 & \underline{0.0273} & 54.27 & 53.41 & 1.2729 & 84.82 & 43.42 & 1.4316 & 37.40 & 31.04 & 5.6593 & 51.43 & 35.68 \\
LvnNE & 0.0225 & 26.59 & 24.31 & 0.0476 & 39.52 & 43.02 & 0.2746 & 84.27 & 30.49 & 0.3972 & 27.89 & 27.59 & 2.2357 & 42.65 & 28.18 \\
ProNE & 0.0398 & 42.88 & 41.25 & 0.0859 & 56.21 & 55.98 & 0.6205 & 84.27 & 43.85 & 0.7429 & 32.95 & 33.34 & 7.3855 & 48.06 & 32.29 \\
SktNE & 1.4635 & 38.09 & 34.91 & 1.2574 & 53.41 & 52.19 & 1.1526 & 84.15 & 33.18 & 2.0193 & 30.65 & 33.09 & 6.1359 & 47.02 & 29.37 \\
SktBANE & 0.0586 & 49.44 & 49.1 & 0.0550 & 56.47 & 56.10 & \underline{0.0475} & 85.15 & 46.71 & \underline{0.0583} & 39.99 & 33.39 & \underline{0.2079} & 50.22 & 32.47 \\ \hline
P-GNN & 1.3910 & 46.50 & 42.22 & 3.5859 & 53.74 & 51.86 & 46.0068 & 84.40 & 35.99 & 66.3064 & 30.33 & 32.81 & \multicolumn{3}{c}{OOM} \\
GraLSP & 181.2625 & 41.50 & 41.48 & 184.8798 & 57.04 & \underline{56.44} & 558.6954 & 85.26 & 46.71 & 187.3734 & 40.93 & \textbf{40.04} & 531.6974 & 55.41 & 32.10 \\
DGI & 1.1951 & 46.31 & 44.06 & 1.6137 & 55.92 & 54.04 & 1.8235 & 85.61 & \textbf{48.98} & 2.2623 & 42.50 & 34.41 & 14.7120 & 56.20 & 32.01 \\
GMAE & 1.2607 & 43.78 & 41.19 & 1.4065 & \underline{57.33} & 56.43 & 0.9423 & \underline{85.72} & 48.39 & 1.9079 & 40.74 & 33.61 & 1.9021 & \underline{59.51} & 36.00 \\
GMAE2 & 2.5425 & 43.50 & 42.40 & 4.3148 & 51.24 & 49.78 & 5.1604 & 85.39 & 44.25 & 6.1243 & 38.26 & 32.10 & 13.8681 & \underline{57.13} & 37.16 \\
GGD & 1.7932 & 46.41 & 39.39 & 1.8773 & 56.14 & 54.59 & 1.9534 & 85.46 & 48.33 & 1.9987 & \underline{43.25} & 34.21 & 3.6673 & 50.24 & 30.36 \\
GECL & 4.5276 & 48.28 & 46.98 & 4.7977 & 55.28 & 54.42 & 5.6982 & \underline{85.70} & 48.36 & 6.7497 & 37.92 & 31.02 & 58.3116 & 50.50 & 29.54 \\ \hline
\textbf{RFA(H)} & \textbf{0.0014} & \textbf{56.66} & \textbf{54.64} & \underline{0.0252} & \underline{57.34} & \underline{56.63} & \textbf{0.0168} & \textbf{85.88} & \underline{48.40} & \underline{0.0263} & \textbf{44.15} & \underline{35.64} & \underline{0.0885} & \textbf{60.03} & \textbf{40.09} \\
\textbf{RFA(L)} & \underline{0.0016} & 26.81 & 13.03 & \textbf{0.0237} & 55.40 & 53.50 & \underline{0.0201} & 84.27 & 30.49 & \textbf{0.0261} & 23.01 & 7.48 & \textbf{0.0850} & 42.88 & 31.57 \\ \hline
\end{tabular}
%\vspace{-0.2cm}
\end{table*}

\begin{table*}[]\scriptsize
\caption{Evaluation results of position embedding in terms of inference time (sec), micro-F1 (\%), and macro-F1 (\%).}\label{Tab:Eva-Pos}
\vspace{-0.2cm}
\begin{tabular}{l|p{0.55cm}p{0.5cm}l|p{0.6cm}p{0.5cm}l|p{0.5cm}p{0.5cm}l|p{0.5cm}p{0.5cm}l|p{0.5cm}p{0.5cm}l}
\hline
\multirow{2}{*}{\textbf{}} & \multicolumn{3}{c|}{\textbf{PPI}} & \multicolumn{3}{c|}{\textbf{BlogCatalog}} & \multicolumn{3}{c|}{\textbf{Flickr}} & \multicolumn{3}{c|}{\textbf{Youtube}} & \multicolumn{3}{c}{\textbf{Orkut}} \\ \cline{2-16} 
 & \tiny{\textbf{Time}$\downarrow$} & \tiny{\textbf{\text{Micro-}}$\uparrow$} & \tiny{\textbf{\text{Macro-F1}}$\uparrow$} & \tiny{\textbf{Time}$\downarrow$} & \tiny{\textbf{\text{Micro-}}$\uparrow$} & \tiny{\textbf{\text{Macro-F1}}$\uparrow$} & \tiny{\textbf{Time}$\downarrow$} & \tiny{\textbf{\text{Micro-}}$\uparrow$} & \tiny{\textbf{\text{Macro-F1}}$\uparrow$} & \tiny{\textbf{Time}$\downarrow$} & \tiny{\textbf{\text{Micro-}}$\uparrow$} & \tiny{\textbf{\text{Macro-F1}}$\uparrow$} & \multicolumn{1}{l}{\tiny{\textbf{Time}$\downarrow$}} & \textbf{\tiny{\text{Micro-}}}$\uparrow$ & \textbf{\tiny{\text{Macro-F1}}}$\uparrow$ \\ \hline
node2vec & 209.9584 & 20.00 & 13.60 & 2081.1264 & 35.63 & 16.14 & \multicolumn{3}{c|}{OOT} & \multicolumn{3}{c|}{OOT} & \multicolumn{3}{c}{OOT} \\
PhN-PPMI & 52.8576 & 20.20 & \underline{16.82} & 616.9198 & \underline{39.09} & \underline{25.05} & \multicolumn{3}{c|}{OOT} & \multicolumn{3}{c|}{OOT} & \multicolumn{3}{c}{OOT} \\
PaCEr(P) & 149.6007 & 18.46 & 12.54 & 2363.3851 & 29.79 & 18.47 & \multicolumn{3}{c|}{OOT} & \multicolumn{3}{c|}{OOT} & \multicolumn{3}{c}{OOT} \\ \hline
struc2vec & 292.9071 & 8.51 & 5.15 & 1671.7530 & 14.95 & 3.87 & \multicolumn{3}{c|}{OOT} & \multicolumn{3}{c|}{OOT} & \multicolumn{3}{c}{OOT} \\
PhN-HK & 350.5650 & 9.25 & 4.25 & 3541.9192 & 17.03 & 3.43 & \multicolumn{3}{c|}{OOT} & \multicolumn{3}{c|}{OOT} & \multicolumn{3}{c}{OOT} \\
PaCEr(I) & 135.4221 & 9.75 & 5.84 & 2269.1116 & 16.34 & 3.48 & \multicolumn{3}{c|}{OOT} & \multicolumn{3}{c|}{OOT} & \multicolumn{3}{c}{OOT} \\ \hline
RandNE & 2.9658 & 14.97 & 9.25 & 55.3924 & 27.33 & 16.18 & 16.2782 & 29.67 & 7.91 & 14.1033 & 34.06 & 28.80 & \multicolumn{1}{l}{86.0596} & \underline{72.03} & \underline{71.98} \\
LvnNE & \underline{0.3773} & 15.13 & 9.01 & 1.8864 & 23.18 & 13.28 & 16.4049 & 31.51 & 19.62 & 46.1556 & 40.31 & 29.17 & \multicolumn{1}{l}{239.8252} & 68.47 & 67.65 \\
ProNE & 1.0387 & \underline{20.75} & \underline{15.38} & 10.7159 & 38.17 & 20.36 & 76.9398 & \underline{39.55} & \underline{23.13} & 100.4127 & \underline{42.27} & \underline{31.30} & \multicolumn{1}{l}{519.6670} & 66.94 & 65.19 \\
SktNE & 1.8351 & \textbf{20.88} & 14.60 & 3.8676 & \textbf{40.23} & \underline{21.85} & 26.9919 & \textbf{40.62} & \underline{25.04} & 37.8038 & \textbf{42.83} & \textbf{32.72} & \multicolumn{1}{l}{430.0041} & \underline{73.80} & \underline{72.84} \\
SktBANE & 0.4171 & 13.56 & 9.71 & \underline{0.4222} & 24.86 & 8.37 & \underline{4.8537} & 25.51 & 6.09 & \underline{1.9685} & 33.44 & 20.59 & \multicolumn{1}{l}{\underline{37.0508}} & 8.26 & 6.33 \\ \hline
P-GNN & 45.6083 & 11.62 & 7.10 & 319.1852 & 20.63 & 6.48 & \multicolumn{3}{c|}{OOM} & \multicolumn{3}{c|}{OOM} & \multicolumn{3}{c}{OOM} \\
GraLSP & 202.0369 & 8.61 & 4.60 & 1879.7471 & 16.89 & 4.26 & \multicolumn{3}{c|}{OOM} & \multicolumn{3}{c|}{OOM} & \multicolumn{3}{c}{OOM} \\
DGI & 4.2957 & 10.98 & 6.17 & 19.2401 & 20.17 & 5.31 & 83.1707 & 20.73 & 2.70 & 92.9302 & 26.41 & 9.62 & \multicolumn{3}{c}{OOM} \\
GMAE & 1.1947 & 9.63 & 5.50 & 1.5246 & 20.57 & 11.26 & 22.1693 & 25.42 & 12.50 & 85.6405 & 32.24 & 19.44 & \multicolumn{3}{c}{OOM} \\
GMAE2 & 14.2143 & 14.56 & 11.66 & 5.8966 & 27.14 & 15.60 & 42.1601 & 33.35 & 19.95 & 249.8887 & 30.61 & 16.04 & \multicolumn{3}{c}{OOM} \\
GGD & 1.9915 & 9.83 & 3.44 & 5.5962 & 16.83 & 3.05 & 88.2213 & 19.34 & 1.92 & 97.3853 & 26.34 & 7.94 & \multicolumn{1}{l}{710.4336} & 4.19 & 1.03 \\
GECL & 4.9530 & 7.97 & 5.09 & 18.4350 & 15.24 & 3.71 & \multicolumn{3}{c|}{OOM} & \multicolumn{3}{c|}{OOM} & \multicolumn{3}{c}{OOM} \\ \hline
\textbf{RFA(L)} & \textbf{0.1777} & \underline{20.38} & \textbf{17.32} & \underline{0.2872} & \underline{38.98} & \textbf{25.22} & \underline{0.5904} & \underline{37.47} & \textbf{27.15} & \textbf{0.4727} & \underline{41.57} & \underline{30.59} & \multicolumn{1}{l}{\underline{3.2474}} & \textbf{77.22} & \textbf{77.55} \\
\textbf{RFA(H)} & \underline{0.2050} & 7.92 & 2.67 & \textbf{0.2860} & 16.58 & 2.55 & \textbf{0.5784} & 16.51 & 0.46 & \underline{0.4791} & 25.35 & 6.08 & \multicolumn{1}{l}{\textbf{2.8112}} & 3.18 & 0.32 \\ \hline
\end{tabular}
%\vspace{-0.2cm}
\end{table*}

\begin{table}[]\tiny
\caption{Detailed inference time (sec)$\downarrow$ of RFA.}\label{Tab:Time}
\vspace{-0.2cm}
\begin{tabular}{c|l|p{0.29cm}p{0.25cm}p{0.35cm}p{0.25cm}p{0.37cm}|p{0.25cm}p{0.33cm}p{0.27cm}p{0.33cm}p{0.39cm}}
\hline
\multicolumn{1}{l|}{} &  & \textbf{Europe} & \textbf{USA} & \textbf{\text{Reality}} & \textbf{Actor} & \textbf{Film} & \textbf{PPI} & \textbf{BlogC} & \textbf{Flickr} & \textbf{Youtube} & \textbf{Orkut} \\ \hline
\multirow{2}{*}{\textbf{RFA(L)}} & CPU & \text{\textbf{0.0016}} & \textbf{0.0237} & \text{\textbf{0.0201}} & \textbf{0.0261} & 2.2006 & 0.5580 & 4.5600 & 60.8398 & 52.6311 & 362.2383 \\
 & GPU & 0.0514 & 0.0966 & 0.0575 & 0.0459 & \textbf{0.0850} & \textbf{0.1777} & \textbf{0.2872} & \textbf{0.5904} & \textbf{0.4727} & \textbf{3.2474} \\ \hline
\multirow{2}{*}{\textbf{RFA(H)}} & CPU & \textbf{0.0014} & \textbf{0.0252} & \textbf{0.0168} & \textbf{0.0263} & 2.2296 & 0.5468 & 4.8438 & 60.0957 & 51.0935 & 225.2535 \\
 & GPU & 0.0610 & 0.0920 & 0.0608 & 0.0640 & \textbf{0.0885} & \textbf{0.2050} & \textbf{0.2860} & \textbf{0.5784} & \textbf{0.4791} & \textbf{2.8112} \\ \hline
\end{tabular}
\vspace{-0.3cm}
\end{table}

The mean evaluation results for identity and position embedding are depicted in Tables~\ref{Tab:Eva-Ide} and \ref{Tab:Eva-Pos}, where a metric is in \textbf{bold} or \underline{underlined} if it performs the best or with the top-$3$.
Table~\ref{Tab:Time} reports the detailed inference time of \textbf{RFA(L)} and \textbf{RFA(H)} ran on CPU and GPU.

In Table~\ref{Tab:Eva-Ide}, \textbf{RFA(H)} achieves much better embedding quality than \textbf{RFA(L)}, consistent with the better quality of identity embedding baselines (e.g., \textit{struc2vec}) than position embedding methods (e.g., \textit{node2vec}). Similarly, the much better quality of \textbf{RFA(L)} than \textbf{RFA(H)} in Table~\ref{Tab:Eva-Pos} is consistent with the better quality of position embedding baselines.
These results validate our investigation (see Fig.~\ref{Fig:Toy} and Appendix~\ref{App:Toy}) that \textit{low- and high-pass information may respectively characterize node identities and positions}.

In Table~\ref{Tab:Time}, the case of running \textbf{RFA} on CPU has better efficiency than that on GPU for small datasets (e.g., \textit{Europe} and \textit{USA}). Whereas, one can achieve much better efficiency on GPU for large datasets (e.g., \textit{Youtube} and \textit{Orkut}).
In each setting, \textbf{RFA(L)} and \textbf{RFA(H)} have close inference time. It implies that the choice of low- and high-pass filters will not significantly affect the efficiency of \textbf{RFA}.

Different from GNN-based methods (e.g., DGI), \textbf{RFA} serves as an extreme ablation study about GNN feature aggregation without any feature extraction, ED, and training.
Surprisingly, \textbf{RFA} can achieve much better embedding quality even with random noise inputs and only one FFP.
In particular, it is unclear for most GNN-based approaches which topology properties (e.g., node identities or positions) they can capture. As a result, these baselines may suffer from poor quality for identity and position embedding, although they were claimed to be powerful in handling tasks on attributed graphs.
Most of them also encounter the OOM and OOT exceptions on large datasets (e.g., \textit{Youtube} and \textit{Orkut}). Whereas, \textbf{RFA} can ensure significantly better efficiency and scalability.

In summary, \textbf{RFA} can always ensure the best inference efficiency, even compared with efficient embedding baselines (e.g., \textit{RandNE} and \textit{SketchNE}). Moreover, \textbf{RFA} can always achieve the top-$3$ embedding quality and even perform the best in most cases.

\begin{figure}[t]
  \centering
  \includegraphics[width=0.6\linewidth, trim=20 22 20 20,clip]{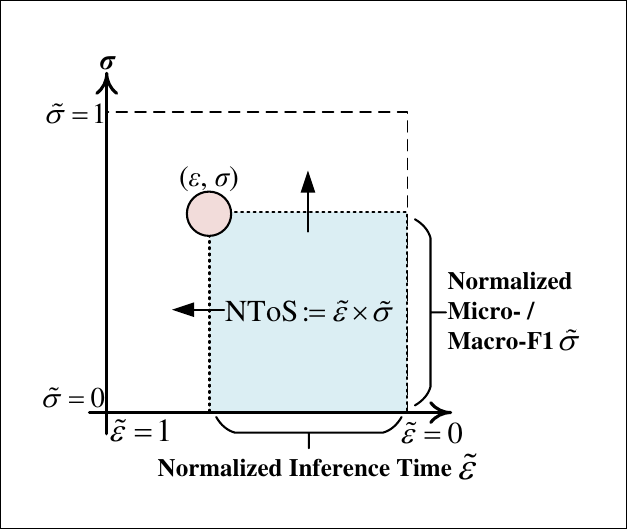}
  \vspace{-0.2cm}
  \caption{Intuition of computing NToS for trade-off analysis.
  }\label{Fig:NToS}
  \vspace{-0.3cm}
\end{figure}

\subsection{Trade-off Analysis}\label{Sec:NTos}

\begin{table*}[]\scriptsize
\caption{Trade-off analysis for identity and position embedding in terms of NToS$\uparrow$.}\label{Tab:NToS}
\vspace{-0.2cm}
\begin{tabular}{l|p{0.35cm}p{0.45cm}|p{0.35cm}p{0.45cm}|p{0.35cm}p{0.45cm}|p{0.35cm}p{0.45cm}|p{0.35cm}p{0.45cm}|p{0.35cm}p{0.45cm}|p{0.35cm}p{0.45cm}|p{0.35cm}p{0.45cm}|p{0.35cm}p{0.45cm}|p{0.35cm}p{0.45cm}}
\hline
\multirow{3}{*}{\textbf{}} & \multicolumn{2}{c|}{\textbf{Europe}} & \multicolumn{2}{c|}{\textbf{USA}} & \multicolumn{2}{c|}{\textbf{Reality-Call}} & \multicolumn{2}{c|}{\textbf{Actor}} & \multicolumn{2}{c|}{\textbf{Film}} & \multicolumn{2}{c|}{\textbf{PPI}} & \multicolumn{2}{c|}{\textbf{BlogCatalog}} & \multicolumn{2}{c|}{\textbf{Flickr}} & \multicolumn{2}{c|}{\textbf{Youtube}} & \multicolumn{2}{c}{\textbf{Orkut}} \\ \cline{2-21} 
 & \tiny{\textbf{Time}~\&} & \tiny{\textbf{Time}~\&} & \tiny{\textbf{Time}~\&} & \tiny{\textbf{Time}~\&} & \tiny{\textbf{Time}~\&} & \tiny{\textbf{Time}~\&} & \tiny{\textbf{Time}~\&} & \tiny{\textbf{Time}~\&} & \tiny{\textbf{Time}~\&} & \tiny{\textbf{Time}~\&} & \tiny{\textbf{Time}~\&} & \tiny{\textbf{Time}~\&} & \tiny{\textbf{Time}~\&} & \tiny{\textbf{Time}~\&} & \tiny{\textbf{Time}~\&} & \tiny{\textbf{Time}~\&} & \tiny{\textbf{Time}~\&} & \tiny{\textbf{Time}~\&} & \multicolumn{1}{l}{\tiny{\textbf{Time}~\&}} & \tiny{\textbf{Time}~\&} \\
 & \tiny{\textbf{\text{Micro}}} & \tiny{\textbf{\text{Macro}}} & \tiny{\textbf{\text{Micro}}} & \tiny{\textbf{\text{Macro}}} & \tiny{\textbf{\text{Micro}}} & \tiny{\textbf{\text{Macro}}} & \tiny{\textbf{\text{Micro}}} & \tiny{\textbf{\text{Macro}}} & \tiny{\textbf{\text{Micro}}} & \tiny{\textbf{\text{Macro}}} & \tiny{\textbf{\text{Micro}}} & \tiny{\textbf{\text{Macro}}} & \tiny{\textbf{\text{Micro}}} & \tiny{\textbf{\text{Macro}}} & \tiny{\textbf{\text{Micro}}} & \tiny{\textbf{\text{Macro}}} & \tiny{\textbf{\text{Micro}}} & \tiny{\textbf{\text{Macro}}} & \multicolumn{1}{l}{\tiny{\textbf{\text{Micro}}}} & \tiny{\textbf{\text{Macro}}} \\ \hline
node2vec & 0.2750 & 0.2846 & 0.5109 & 0.4257 & 0.3115 & 0.1240 & 0.0000 & 0.2065 & 0.1223 & 0.1276 & 0.3739 & 0.2937 & 0.3374 & 0.2435 & \multicolumn{2}{c|}{OOT} & \multicolumn{2}{c|}{OOT} & \multicolumn{2}{c}{OOT} \\
PhN-PPMI & 0.1772 & 0.2128 & 0.5344 & 0.4007 & 0.4356 & 0.3899 & 0.2554 & 0.0000 & 0.1001 & 0.2299 & 0.8049 & 0.8190 & 0.7886 & 0.8196 & \multicolumn{2}{c|}{OOT} & \multicolumn{2}{c|}{OOT} & \multicolumn{2}{c}{OOT} \\
PaCEr(P) & 0.6557 & 0.7234 & 0.7281 & 0.6697 & 0.0000 & 0.6372 & 0.4969 & 0.3345 & 0.0000 & 0.0000 & 0.4660 & 0.3760 & 0.1953 & 0.2314 & \multicolumn{2}{c|}{OOT} & \multicolumn{2}{c|}{OOT} & \multicolumn{2}{c}{OOT} \\ \hline
struc2vec & \underline{0.8194} & \underline{0.8690} & 0.6544 & 0.6372 & 0.5160 & 0.7309 & 0.4745 & 0.3266 & 0.1648 & 0.2318 & 0.0069 & 0.0203 & 0.0000 & 0.0195 & \multicolumn{2}{c|}{OOT} & \multicolumn{2}{c|}{OOT} & \multicolumn{2}{c}{OOT} \\
PhN-HK & 0.7870 & 0.7825 & 0.8430 & 0.8430 & 0.0000 & 0.0000 & 0.0000 & 0.0000 & 0.0199 & 0.0241 & 0.0000 & 0.0000 & 0.0000 & 0.0000 & \multicolumn{2}{c|}{OOT} & \multicolumn{2}{c|}{OOT} & \multicolumn{2}{c}{OOT} \\
PaCEr(I) & \underline{0.9345} & \underline{0.9801} & 0.8182 & 0.7534 & 0.7115 & 0.6440 & 0.6255 & 0.3359 & 0.0413 & 0.0708 & 0.0847 & 0.1062 & 0.0198 & 0.0070 & \multicolumn{2}{c|}{OOT} & \multicolumn{2}{c|}{OOT} & \multicolumn{2}{c}{OOT} \\ \hline
RandNE & 0.8377 & 0.8674 & 0.7607 & 0.6945 & 0.6385 & 0.6982 & 0.5912 & 0.2854 & 0.5047 & 0.6292 & 0.5379 & 0.4153 & 0.4821 & 0.5830 & 0.3985 & 0.1949 & 0.4426 & \underline{0.7958} & \multicolumn{1}{l}{\underline{0.8202}} & \underline{0.8186} \\
LvnNE & 0.0000 & 0.0000 & 0.0000 & 0.0000 & 0.4522 & 0.0000 & 0.0169 & 0.0119 & 0.0000 & 0.0000 & 0.5543 & 0.4011 & 0.3254 & 0.4612 & \underline{0.4687} & \underline{0.5749} & \underline{0.6920} & 0.6998 & \multicolumn{1}{l}{\underline{0.5857}} & \underline{0.5794} \\
ProNE & 0.5416 & 0.5584 & 0.8605 & \underline{0.8660} & 0.4520 & 0.7220 & 0.3227 & 0.4680 & 0.3109 & 0.3447 & \underline{0.9875} & \underline{0.8581} & \underline{0.9158} & \underline{0.7785} & 0.1223 & 0.1082 & 0.5790 & 0.5650 & \multicolumn{1}{l}{0.2318} & 0.2262 \\
SktNE & 0.3794 & 0.3467 & 0.7116 & 0.6089 & 0.4110 & 0.1453 & 0.1835 & 0.4477 & 0.2512 & 0.0998 & \textbf{0.9953} & \underline{0.8002} & \textbf{0.9990} & \underline{0.8471} & \underline{0.6987} & \underline{0.6403} & \underline{0.8503} & \underline{0.8503} & \multicolumn{1}{l}{0.3780} & 0.3721 \\
SktBANE & 0.7597 & 0.8171 & \underline{0.8740} & 0.8742 & 0.7517 & 0.8772 & 0.7485 & 0.4722 & 0.4356 & 0.3602 & 0.4327 & 0.4514 & 0.3920 & 0.2400 & 0.2758 & 0.1572 & 0.4280 & 0.5074 & \multicolumn{1}{l}{0.0531} & 0.0660 \\ \hline
P-GNN & 0.6570 & 0.5860 & 0.7192 & 0.5795 & 0.4681 & 0.2804 & 0.1553 & 0.4024 & \multicolumn{2}{c|}{OOM} & 0.2461 & 0.2295 & 0.2045 & 0.1408 & \multicolumn{2}{c|}{OOM} & \multicolumn{2}{c|}{OOM} & \multicolumn{2}{c}{OOM} \\
GraLSP & 0.0000 & 0.0000 & 0.0000 & 0.0000 & 0.2386 & 0.2653 & 0.6780 & \textbf{0.8419} & 0.6710 & 0.3008 & 0.0210 & 0.0354 & 0.0360 & 0.0256 & \multicolumn{2}{c|}{OOM} & \multicolumn{2}{c|}{OOM} & \multicolumn{2}{c}{OOM} \\
DGI & 0.6515 & 0.6469 & 0.8385 & 0.7303 & 0.9061 & \textbf{0.9977} & \underline{0.8985} & 0.5521 & 0.7778 & 0.3208 & 0.2305 & 0.1944 & 0.2054 & 0.1014 & 0.0038 & 0.0018 & 0.0027 & 0.0427 & \multicolumn{2}{c}{OOM} \\
GMAE & 0.5677 & 0.5527 & \underline{0.9117} & \underline{0.8897} & \underline{0.9445} & \underline{0.9670} & 0.7926 & 0.4889 & \underline{0.9698} & \underline{0.6564} & 0.1282 & 0.1480 & 0.2222 & 0.3702 & 0.2154 & 0.3161 & 0.2356 & 0.3056 & \multicolumn{2}{c}{OOM} \\
GMAE2 & 0.5545 & 0.5881 & 0.5904 & 0.4414 & 0.8280 & 0.7394 & 0.6406 & 0.3679 & \underline{0.8313} & \underline{0.7523} & 0.4900 & 0.5685 & 0.4814 & 0.5652 & 0.3461 & 0.3756 & 0.0000 & 0.0000 & \multicolumn{2}{c}{OOM} \\
GGD & 0.6526 & 0.4923 & 0.8486 & 0.7656 & 0.8551 & 0.9625 & \underline{0.9440} & \underline{0.5364} & 0.4365 & 0.1829 & 0.1433 & 0.0000 & 0.0743 & 0.0000 & 0.0000 & 0.0000 & 0.0000 & 0.0000 & \multicolumn{1}{l}{0.0000} & 0.0000 \\
GECL & 0.7033 & 0.7288 & 0.7918 & 0.7424 & \underline{0.9321} & 0.9596 & 0.6198 & 0.2825 & 0.4474 & 0.1131 & 0.0000 & 0.1173 & 0.0114 & 0.0296 & \multicolumn{2}{c|}{OOM} & \multicolumn{2}{c|}{OOM} & \multicolumn{2}{c}{OOM} \\ \hline
\textbf{RFA} & \textbf{1.0000} & \textbf{1.0000} & \textbf{0.9190} & \textbf{0.9098} & \textbf{1.0000} & \underline{0.9686} & \textbf{1.0000} & \underline{0.6508} & \textbf{1.0000} & \textbf{1.0000} & \underline{0.9613} & \textbf{1.0000} & \underline{0.9506} & \textbf{1.0000} & \textbf{0.8520} & \textbf{1.0000} & \textbf{0.9236} & \textbf{0.9140} & \multicolumn{1}{l}{\textbf{1.0000}} & \textbf{1.0000} \\ \hline
\end{tabular}
%\vspace{-0.3cm}
\end{table*}

Based on our evaluation results depicted in Tables~\ref{Tab:Eva-Ide} and \ref{Tab:Eva-Pos}, we follow \cite{qin2023towards} to quantitatively measure the trade-off between inference quality and efficiency via a normalized trade-off score (NToS). Fig.~\ref{Fig:NToS} illustrates the intuition of computing NToS.
Given a quality metric $\sigma$ (e.g., micro- or macro-F1) and an efficiency metric $\varepsilon$ (i.e., inference time) of a method, we can map this method to a point $(\varepsilon, \sigma)$ in a 2D space (e.g., red circle in Fig.~\ref{Fig:NToS}).
\textit{A point close to the top left corner of the 2D space indicates a better trade-off between quality and efficiency}.
Following this intuition, we first normalize $\varepsilon \in (0, +\infty )$ and $\sigma \in [0, 1]$ to $[0, 1]$ via $\tilde \varepsilon : = ({\varepsilon _{\max }} - \varepsilon )/({\varepsilon _{\max }} - {\varepsilon _{\min }})$ and $\tilde \sigma : = (\sigma  - {\sigma _{\min }})/({\sigma _{\max }} - {\sigma _{\min }})$, where  $\varepsilon_{\min}$ and $\varepsilon_{\max}$ denote the minimum and maximum efficiency metrics among all methods; $\sigma_{\min}$ and $\sigma_{\max}$ are the minimum and maximum quality metrics. This normalization step eliminates the magnitude difference between $\sigma$ and $\varepsilon$. For each method, we define ${\mathop{\rm NToS}\nolimits} (\varepsilon ,\sigma ): = \tilde \varepsilon  \times \tilde \sigma$ (i.e., area covered by the blue rectangle in Fig.~\ref{Fig:NToS}), which is within the range $[0, 1]$. Hence, \textit{a larger NoTS (i.e., a larger area of the induced rectangle) indicates a better trade-off between quality and efficiency}.

The trade-off analysis results based on Tables~\ref{Tab:Eva-Ide} and \ref{Tab:Eva-Pos} are shown in Table~\ref{Tab:NToS}, where an NToS metric is in \textbf{bold} or \underline{underlined} if it performs the best or within top-$3$. According to our trade-off analysis, \textbf{RFA} can achieve the best NToS in most cases and always performs within the top-$3$. Therefore, we believe that the proposed \textbf{RFA} method can ensure a better trade-off between quality and efficiency for identity and position embedding over various baselines.

\subsection{Scalability Analysis}
\begin{figure}[t]
  \centering
  \includegraphics[width=0.9\linewidth, trim=25 0 40 5,clip]{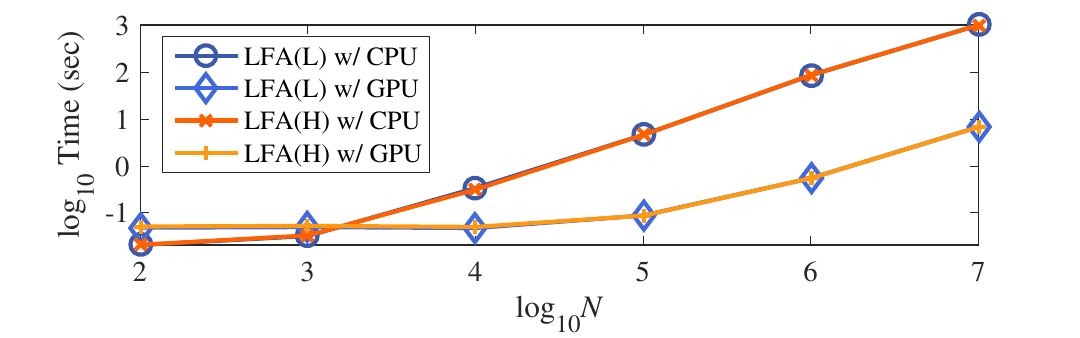}
  \vspace{-0.3cm}
  \caption{Scalability analysis results of RFA.
  }\label{Fig:Sbl}
  \vspace{-0.3cm}
\end{figure}

We further tested the scalability of \textbf{RFA} by applying the two variants to Erdős-Rényi graphs \cite{erdds1959random,batagelj2005efficient}. Concretely, we set a constant average degree $\bar d = 10$ and increased the number of nodes $N$ by setting $N \in \{ 10^2, 10^3, 10^4, 10^5, 10^6, 10^7\}$, which corresponds to setting the average number of edges $\bar M \in \{ 2 \times 10^3, 2 \times 10^4, \cdots, 2 \times 10^8 \}$. Consistent with the parameter settings in Appendix~\ref{App:Exp-Set}, we set $(d, \tau, K) = (64, 20, 10)$ for both \textbf{RFA(L)} and \textbf{RFA(H)}. $\tanh ( \cdot )$ and $\exp ( \cdot )$ were used as the activation functions of \textbf{RFA(L)} and \textbf{RFA(H)}. Z-score normalization was also applied to both variants.
For each setting of $N$ (or $\bar M$), the evaluation procedure was repeated $5$ times, with the mean inference time reported.
Fig.~\ref{Fig:Sbl} visualizes our results of scalability analysis, where two variants of \textbf{RFA} have almost the same inference time. Consistent with Table~\ref{Tab:Time}, the CPU versions of \textbf{RFA(L)} and \textbf{RFA(H)} have slightly faster inference time than the corresponding GPU versions when $N$ (or $\bar M$) is small (e.g., $N \le 10^3$ w.r.t. $\bar M \le 2 \times 10^4$). Whereas, the GPU versions have significantly better efficiency for $N \ge 10^4$ ($\bar M \ge 2 \times 10^5$).
In particular, \textit{the inference time of running \textbf{RFA} on CPU scales linearly with the increase of $N$ (or $\bar M$)}. Moreover, \textit{the GPU versions also have similar variation tendencies regarding inference time when $N \ge 10^5$ ($\bar M \ge 2 \times 10^6$)}.

\subsection{Parameter Analysis \& Ablation Study}

\begin{figure*}[]
 \begin{minipage}{0.24\linewidth}
 \subfigure[RFA(L), PPI]{
  \includegraphics[width=\textwidth,trim=50 0 45 5,clip]{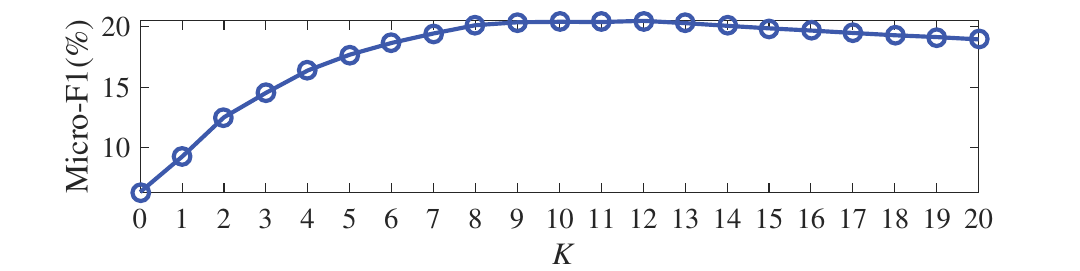}
  }
 \end{minipage}
 \begin{minipage}{0.24\linewidth}
 \subfigure[RFA(L), Youtube]{
  \includegraphics[width=\textwidth,trim=50 0 45 5,clip]{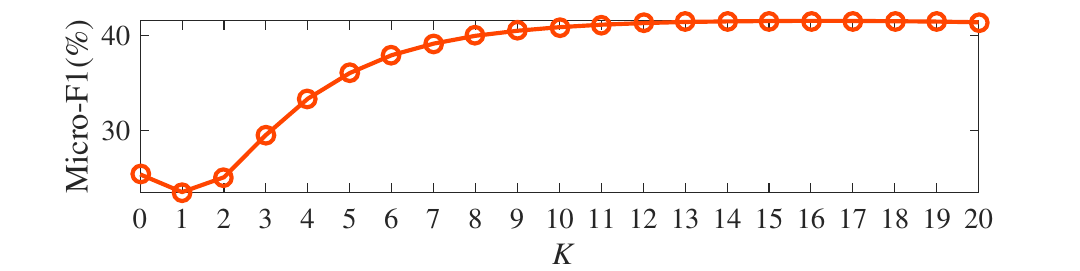}
  }
 \end{minipage}
 %----------
 \begin{minipage}{0.24\linewidth}
 \subfigure[RFA(H), Europe]{
  \includegraphics[width=\textwidth,trim=50 0 45 5,clip]{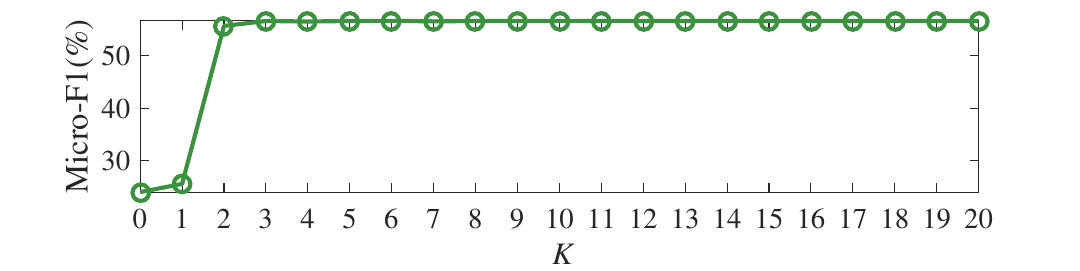}
  }
 \end{minipage}
 \begin{minipage}{0.24\linewidth}
 \subfigure[RFA(H), Actor]{
  \includegraphics[width=\textwidth,trim=50 0 45 5,clip]{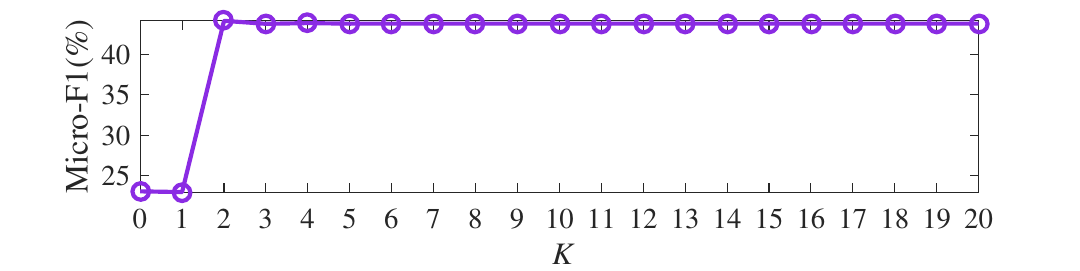}
  }
 \end{minipage}
 %----------
 \vspace{-0.4cm}
 \caption{Parameter analysis w.r.t. $K$ on PPI, Youtube, Europe, and Actor in terms of micro-F1(\%).
 }\label{Fig:Param-K}
 %\vspace{-0.2cm}
\end{figure*}

\begin{figure*}[]
 \begin{minipage}{0.24\linewidth}
 \subfigure[RFA(L), PPI]{
  \includegraphics[width=\textwidth,trim=40 0 42 5,clip]{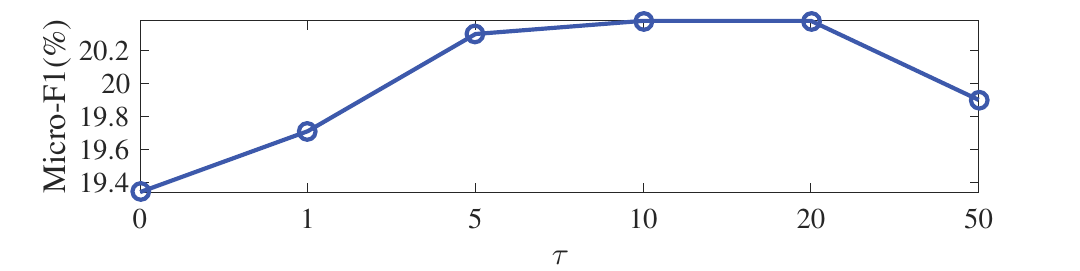}
  }
 \end{minipage}
 \begin{minipage}{0.24\linewidth}
 \subfigure[RFA(L), Youtube]{
  \includegraphics[width=\textwidth,trim=38 0 42 5,clip]{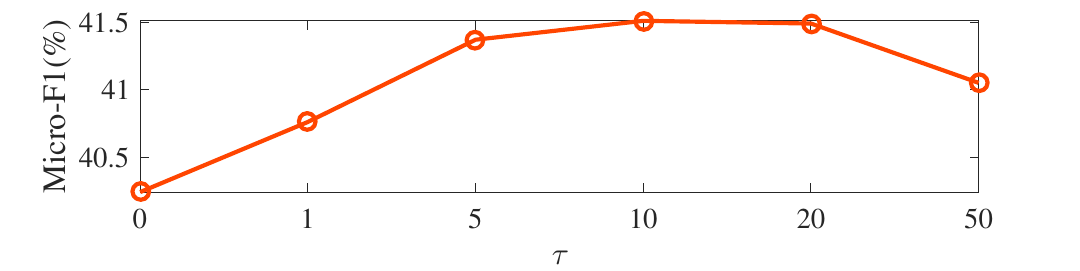}
  }
 \end{minipage}
 %----------
 \begin{minipage}{0.24\linewidth}
 \subfigure[RFA(H), Europe]{
  \includegraphics[width=\textwidth,trim=43 0 42 5,clip]{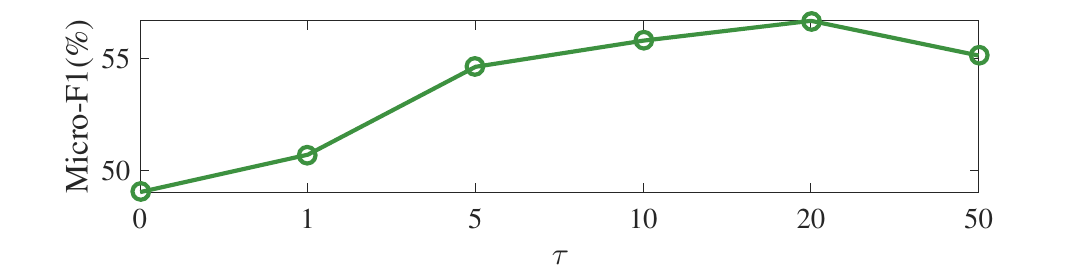}
  }
 \end{minipage}
 \begin{minipage}{0.24\linewidth}
 \subfigure[RFA(H), Actor]{
  \includegraphics[width=\textwidth,trim=43 0 42 5,clip]{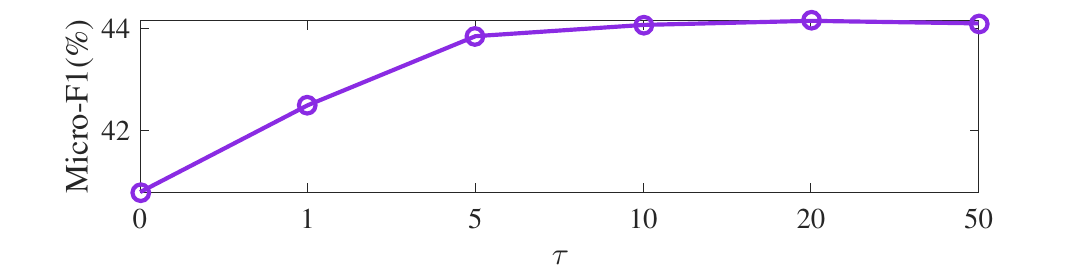}
  }
 \end{minipage}
 %----------
 \vspace{-0.4cm}
 \caption{Parameter analysis w.r.t. $\tau$ on PPI, Youtube, Europe, and Actor in terms of micro-F1(\%).
 }\label{Fig:Param-tau}
 %\vspace{-0.4cm}
\end{figure*}

\begin{table}[]\scriptsize
\caption{Ablation study of $f_{\rm{act}}(\cdot)$ and normalization.}\label{Tab:Abl}
\vspace{-0.2cm}
\begin{tabular}{c|c|p{1cm}|p{0.3cm}p{0.3cm}p{0.55cm}|p{0.3cm}p{0.35cm}p{0.5cm}}
\hline
\multicolumn{1}{l|}{\textbf{Methods}} & \multicolumn{1}{l|}{\textbf{Datasets}} &  & tanh & exp & \tiny{w/o~$f_{\rm{act}}$} & \tiny{\text{z-norm}} & \tiny{\text{$l2$-norm}} & \tiny{\text{w/o~norm}} \\ \hline
\multirow{4}{*}{\textbf{RFA(L)}} & \multirow{2}{*}{\textbf{PPI}} & \text{Micro-F1}(\%) & \textbf{20.38} & 11.67 & 19.91 & \textbf{20.38} & 12.25 & 6.43 \\
 &  & \text{Macro-F1}(\%) & \textbf{17.32} & 7.56 & 15.47 & \textbf{17.32} & 5.37 & 1.63 \\ \cline{2-9} 
 & \multirow{2}{*}{\textbf{Youtube}} & \text{Micro-F1}(\%) & \textbf{41.57} & 25.35 & 41.51 & 36.20 & \textbf{41.57} & 25.35 \\
 &  & \text{Macro-F1}(\%) & \textbf{30.59} & 6.08 & 30.54 & 28.01 & \textbf{30.59} & 6.08 \\ \hline
\multirow{4}{*}{\textbf{RFA(H)}} & \multirow{2}{*}{\textbf{Europe}} & \text{Micro-F1}(\%) & 26.78 & \textbf{56.66} & 26.78 & 40.09 & 25.78 & \textbf{56.66} \\
 &  & \text{Macro-F1}(\%) & 13.01 & \textbf{54.64} & 13.01 & 39.70 & 12.18 & \textbf{54.64} \\ \cline{2-9} 
 & \multirow{2}{*}{\textbf{Actor}} & \text{Micro-F1}(\%) & 23.01 & \textbf{44.15} & 23.01 & 28.43 & 23.01 & \textbf{44.15} \\
 &  & \text{Macro-F1}(\%) & 7.48 & \textbf{35.64} & 7.48 & 25.17 & 7.48 & \textbf{35.64} \\ \hline
\end{tabular}
%\vspace{-0.4cm}
\end{table}

We also verified the effects of hyper-parameters $\{ K, \tau\}$ as well as settings of activation function $f_{\rm{act}} (\cdot)$ and normalization.
PPI and Youtube w.r.t. position embedding were used as example datasets for the analysis of \textbf{RFA(L)}. Europe and Actor w.r.t. identity embedding were adopted for \textbf{RFA(H)}.
Example parameter analysis results for (\romannumeral1) the number of iterations (i.e., GNN layers) $K$ and (\romannumeral2) degree correction $\tau$ in terms of micro-F1 are shown in Fig.~\ref{Fig:Param-K} and \ref{Fig:Param-tau}, where we adjusted $K \in \{ 0, 1, 2, \cdots, 20\}$ and $\tau \in \{ 0, 1, 10, 20, 50\}$.

With the increase of $K$, the embedding quality of \textbf{RFA(L)} and \textbf{RFA(H)} significantly improves. \textbf{RFA} may achieve the best embedding quality with a relatively small $K$ (e.g., $K < 10$).
Compared with $\tau=0$, the incorporation of degree correction (i.e., $\tau > 0$) can significantly improve the quality of both identity and position embeddings in some cases. However, a large setting of $\tau$ (e.g., $\tau > 50$) may further cause quality degradation.

For $f_{\rm{act}} (\cdot)$, we considered settings of (\romannumeral1) $\tanh ( \cdot )$, (\romannumeral2) $\exp ( \cdot )$, and (\romannumeral3) without non-linear activation. For normalization, we tested cases with (\romannumeral1) $l2$, (\romannumeral2) z-score, and (\romannumeral3) without normalization. Ablation study results regarding $f_{\rm{act}}(\cdot)$ and normalization in terms of micro- and macro-F1 scores are depicted in Table~\ref{Tab:Abl}.
The application of non-linear activation may significantly improve the embedding quality of \textbf{RFA}. Concretely, \textbf{RFA(H)} and \textbf{RFA(L)} prefer $\exp ( \cdot )$ and $\tanh ( \cdot )$ for identity and position embedding, respectively.
Furthermore, normalization may also help improve the embedding quality. The optimal choice of $f_{\rm{act}} (\cdot)$ may differ in terms of (\romannumeral1) concrete settings of \textbf{RFA} and (\romannumeral2) datasets. In some cases, \textbf{RFA} without non-linear activation is enough to ensure high embedding quality.

\section{Related Work}\label{Sec:Rel}

\textbf{Identity \& Position Graph Embedding}.
Rossi et al. \cite{rossi2020proximity} provided an overview covering representative methods for the two types of embedding (e.g., (\romannumeral1) \textit{struc2vec} \cite{ribeiro2017struc2vec} and \textit{GraphWave} \cite{donnat2018learning} for identity embedding; (\romannumeral2) \textit{DeepWalk} \cite{perozzi2014deepwalk} and \textit{node2vec} \cite{grover2016node2vec} for position embedding).
In particular, there have been studies exploring the correlation between node identities and positions. Zhu et al. \cite{zhu2021node} introduced a \textit{PhUSION} framework involving three steps and demonstrated that different settings of these steps may derive different types of embeddings. Based on the invariant theory, Srinivasan et al. \cite{srinivasan2019equivalence} proved that the relation between identity and position embeddings is analogous to that between a probability distribution and its samples.
\textit{IRWE} \cite{qin2024irwe} jointly learns the two types of embeddings following hypotheses that (\romannumeral1) AWs and RWs can respectively capture node identities and positions; (\romannumeral2) nodes with different identities may have different contributions in forming a community.
Yan et al. \cite{yan2024pacer} developed \textit{PaCEr}, which first learns position embeddings based on RW with restart proximity distribution and then obtains identity embeddings by sorting the distribution.

However, most of these methods rely heavily on several time-consuming procedures about graph topology (e.g., RW sampling). Few of them explore the correlation between the two types of properties from a GSP perspective.

\textbf{Efficient \& Scalable Graph Embedding}.
To ensure high inference efficiency and scalability, some embedding approaches (e.g., \textit{RandNE} \cite{zhang2018billion} and \textit{SketchBANE} \cite{wu2024time}) applied the Gaussian random projection \cite{arriaga2006algorithmic}, a fast dimension reduction technique that can preserve geometry structures of inputs with rigorous guarantees, to high-dimensional graph features (e.g., derived from ${\bf{A}}$). Following \cite{qiu2018network} that formulates classic embedding methods as MF problems, \textit{NetSMF} \cite{qiu2019netsmf} derives embeddings by applying randomized singular value decomposition (SVD) to an approximated sparse MF objective. \textit{ProNE} \cite{zhang2019prone} enhances the embedding quality of \textit{NetSMF} via a spectral propagation procedure. \textit{LightNE} \cite{qiu2021lightne} optimizes some key steps (e.g., randomized SVD and spectral propagation) of \textit{ProNE} to achieve better efficiency and scalability. \textit{SketchNE} \cite{xie2023sketchne} handles the limitation of \textit{NetSMF} in balancing quality and scalability via fast ED and sparse-sign randomized single-pass SVD.

Nevertheless, most of these approaches are either (\romannumeral1) able to derive only one type of embeddings or (\romannumeral2) unclear to capture which properties (i.e., node identities or positions). Some of them may also suffer from low inference quality caused by the information loss of fast approximation.

\textbf{Spectral-Based GNNs}.
Bo et al. \cite{bo2023survey} gave a survey about spectral-based GNNs.
Related studies aimed to (\romannumeral1) alleviate over-smoothing of conventional GNNs \cite{dong2021adagnn,bo2021beyond,zhu2021simple}, (\romannumeral2) handle the heterophily between graph topology and labels \cite{bo2021beyond,duanunifying}, as well as (\romannumeral3) design better graph convolution kernels \cite{xu2019graph,zhu2021simple,wang2022powerful}.
Different from our problem statement in Section~\ref{Sec:Prob}, most of these techniques rely on some time-consuming \textit{supervised} training procedures (e.g., node classification) designed for \textit{attributed graphs} and seldom explore the roles of different filters in capturing pure topology properties (i.e., node identities or positions).

In contrast to all the aforementioned methods, \textbf{RFA} can achieve a better trade-off between quality and efficiency for \textit{unsupervised identity and position embedding}. It verifies our GSP-based investigation that \textit{high- and low-frequency information in the graph spectral domain can respectively capture node identities and positions}.

\section{Conclusions}\label{Sec:Con}
In this paper, we considered the unsupervised graph embedding and explored the potential of GNN feature aggregation to capture pure topology properties. From a view of GSP, we found that \textit{high- and low-frequency information in the graph spectral domain may characterize node identities and positions, respectively}. Based on this investigation, we proposed RFA, a simple yet effective method for efficient identity and position embedding. It serves as an extreme ablation study regarding GNN feature aggregation, where we (\romannumeral1) used a spectral-based GNN without learnable model parameters as the backbone of RFA, (\romannumeral2) only fed random noises to this backbone, and (\romannumeral3) derived embeddings via just one FFP (i.e., without additional feature extraction, ED, and training).
Inspired by DCSC, we also introduced a degree correction mechanism to each GNN layer, which was further combined with non-linear activation and normalization.
Surprisingly, our experiments demonstrated that even with random inputs and one FFP, two variants of RFA based on high- and low-pass filters can still derive informative identity and position embeddings, which verified our intuitive GSP-based investigation (e.g., Fig.~\ref{Fig:Toy} and Appendix~\ref{App:Toy}). Furthermore, RFA is also efficient and thus can achieve a better trade-off between quality and efficiency for both identity and position embedding over various baselines.
Some possible future research directions of this study are summarized as follows.

\textit{\textbf{Theoretical Guarantees}}.
This study empirically validates the capacities of high- and low-frequency information to characterize node identities and positions. As described in Section~\ref{Sec:Prob}, one can quantitatively measure the two types of properties using statistics about graph topology (e.g., similarity between ego-nets and overlaps of local neighbors).
In our future research, we plan to explore rigorous guarantees of RFA to capture specific topology properties based on existing theoretical results, e.g., the WL kernel to measure similarities between (sub)graphs \cite{shervashidze2011weisfeiler}; AW to reconstruct ego-nets \cite{micali2016reconstructing}; stochastic block models \cite{karrer2011stochastic} and graph-cut minimization objectives \cite{von2007tutorial} to formulate community structures.

\textit{\textbf{Combination with Existing Methods \& Automatic Model Configuration}}.
We believe that this study can provide new insights to design more powerful and efficient GNN architectures to capture pure topology properties.
In our future work, we will use RFA as the backbone of existing GNN-based methods \cite{velivckovicdeep,hou2022graphmae,hou2023graphmae2,zheng2022rethinking} and test whether it can achieve better inference quality for some other tasks (e.g., link prediction) on different types of graphs.
Moreover, $K$ (i.e., the number of GNN layers or iterations) and $\tau$ (i.e., the degree correction term) are set to be hyper-parameters of RFA in this paper. By combining the training objectives of existing approaches, it is possible to develop an automatic configuration scheme. For instance, we can treat $\tau$ as a learnable model parameter, where different layers and even nodes may be automatically assigned different $\tau$ via a training procedure. Instead of manually setting $K$, one can also combine skip connections with the attention mechanism to automatically give weights for different GNN layers.

\textit{\textbf{Extension to Attributed Graphs}}.
As stated in Section~\ref{Sec:Prob}, we considered unsupervised graph embedding with topology as the only available input, due to the complicated correlation between graph topology and attributes \cite{newman2016structure,qin2018adaptive,wang2020gcn,qin2021dual}.
We intend to extend RFA to attributed graphs based on an adaptive integration mechanism. Concretely, when attributes match well with a specific type of topology properties, one can fully utilize the complementary information from attributes to achieve better inference quality. In contrast, when attributes are inconsistent with topology properties, we need to adaptively control the contribution of attributes w.r.t. the inconsistency degree to avoid possible quality degradation.

\begin{acks}
Meng Qin was supported by the Pengcheng Laboratory under the Interdisciplinary Frontier Research project (2025QYB015). Jiahong Liu and Irwin King were partly supported by the Research Grants Council of the Hong Kong Special Administrative Region, China (CUHK 2410072, RGCR1015-23).
\end{acks}


\bibliographystyle{ACM-Reference-Format}
%\bibliography{RFA}

\appendix

\begin{table*}[]\footnotesize
\caption{Parameter settings of RFA(L).}\label{Tab:Parm-L}
\vspace{-0.2cm}
\begin{tabular}{lll|lll|lll|lll|lll}
\hline
\multicolumn{3}{c|}{\textbf{PPI}} & \multicolumn{3}{c|}{\textbf{BlogCatalog}} & \multicolumn{3}{c|}{\textbf{Flickr}} & \multicolumn{3}{c|}{\textbf{Youtube}} & \multicolumn{3}{c}{\textbf{Orkut}} \\ \hline
($d$, $\tau$, $K$) & $f_{\rm{act}}(\cdot)$ & Norm & ($d$, $\tau$, $K$) & $f_{\rm{act}}(\cdot)$ & Norm & ($d$, $\tau$, $K$) & $f_{\rm{act}}(\cdot)$ & norm & ($d$, $\tau$, $K$) & $f_{\rm{act}}(\cdot)$ & Norm & ($d$, $\tau$, $K$) & $f_{\rm{act}}(\cdot)$ & Norm \\ \hline
(256, 20, 10) & tanh & z-norm & (512, 0, 9) & tanh & z-norm & (512, 1, 7) & tanh & z-norm & (128, 10, 14) & tanh & $l2$-norm & (64, 20, 8) & tanh & z-norm \\ \hline
\end{tabular}
%\vspace{-0.2cm}
\end{table*}

\begin{table*}[]\footnotesize
\caption{Parameter settings of RFA(H).}\label{Tab:Parm-H}
\vspace{-0.2cm}
\begin{tabular}{lll|lll|lll|lll|lll}
\hline
\multicolumn{3}{c|}{\textbf{Europe}} & \multicolumn{3}{c|}{\textbf{USA}} & \multicolumn{3}{c|}{\textbf{Reality-Call}} & \multicolumn{3}{c|}{\textbf{Actor}} & \multicolumn{3}{c}{\textbf{Film}} \\ \hline
($d$, $\tau$, $K$) & $f_{\rm{act}}(\cdot)$ & Norm & ($d$, $\tau$, $K$) & $f_{\rm{act}}(\cdot)$ & Norm & ($d$, $\tau$, $K$) & $f_{\rm{act}}(\cdot)$ & Norm & ($d$, $\tau$, $K$) & $f_{\rm{act}}(\cdot)$ & Norm & ($d$, $\tau$, $K$) & $f_{\rm{act}}(\cdot)$ & Norm \\ \hline
(64, 20, 3) & exp & z-norm & (64, 20, 7) & exp & w/o & (128, 20, 2) & exp & w/o & (128, 20, 2) & exp & w/o & (256, 10, 12) & exp & z-norm \\ \hline
\end{tabular}
%\vspace{-0.2cm}
\end{table*}

\begin{figure}[t]
  \centering
  \includegraphics[width=0.75\linewidth, trim=0 0 30 20,clip]{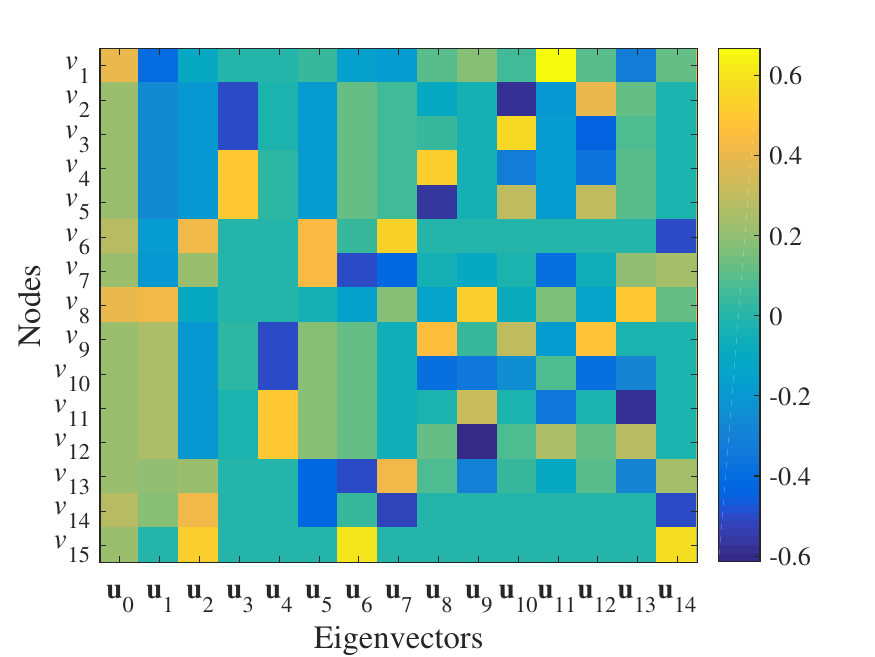}
  \vspace{-0.2cm}
  \caption{Full eigenvectors $\{ {\bf{u}}_r \}$ w.r.t. the normalized graph Laplacian ${\bf{L}}$ of the example graph in Fig.~\ref{Fig:Toy}.
  }\label{Fig:Toy-1-vecs}
  \vspace{-0.6cm}
\end{figure}

\begin{figure}[]
 \begin{minipage}{\linewidth}
 \subfigure[\textbf{Topology and ED on graph Laplacian}]{
  \includegraphics[width=\textwidth,trim=18 30 20 22,clip]{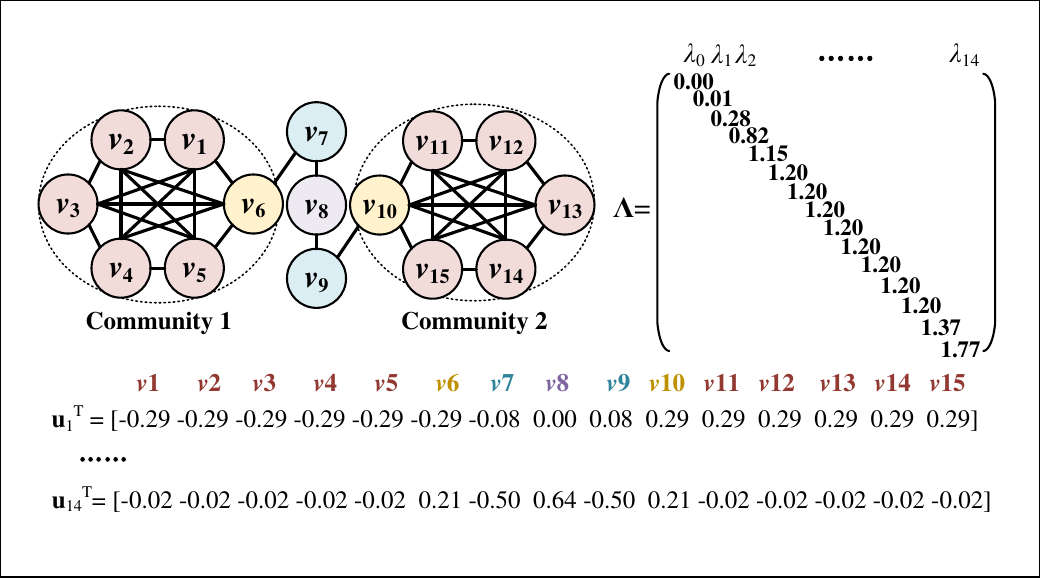}
  }
 \end{minipage}
 \begin{minipage}{0.75\linewidth}
 \subfigure[\textbf{Full eigenvectors}]{
  \includegraphics[width=\textwidth,trim=0 0 30 20,clip]{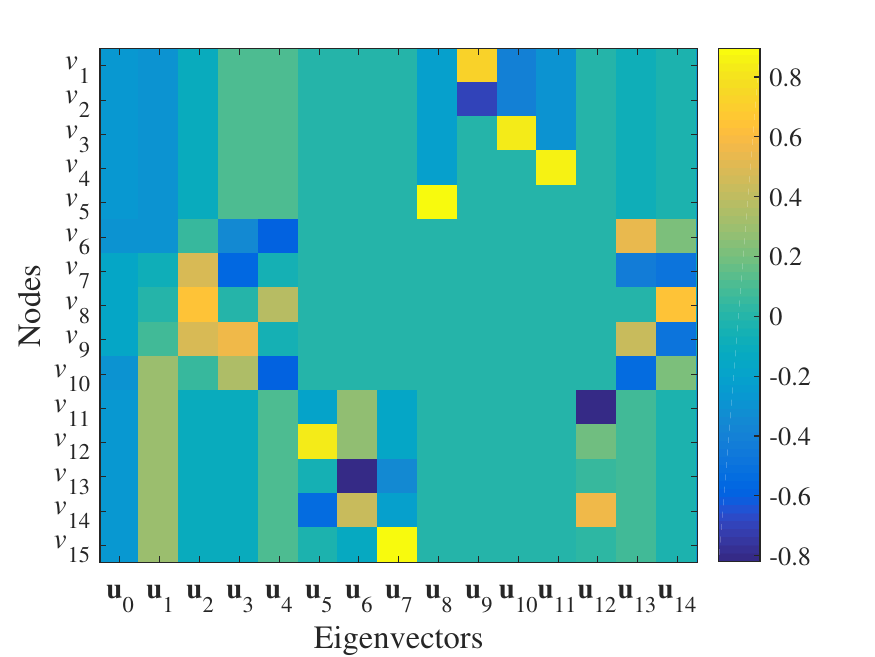}
  }
 \end{minipage}
 %----------
 \vspace{-0.4cm}
 \caption{A further example of node identities and positions as well as the ED of graph Laplacian on the barbell graph, where each color denotes a unique identity; nodes in the same community have similar positions.
 }\label{Fig:Toy-2}
 \vspace{-0.2cm}
\end{figure}

\begin{figure*}[t]
 \begin{minipage}{0.19\linewidth}
 \subfigure[$\tau = 0$]{
  \includegraphics[width=\textwidth,trim=22 22 45 32,clip]{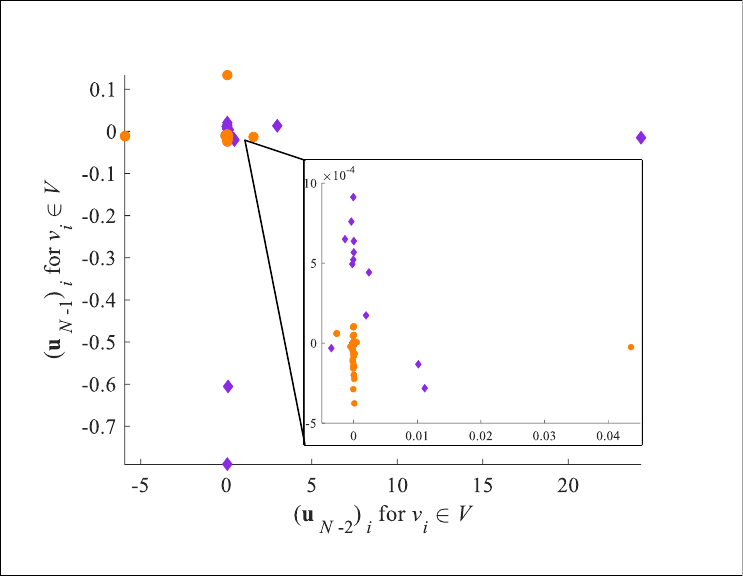}
  }
 \end{minipage}
 \begin{minipage}{0.19\linewidth}
 \subfigure[$\tau = 10$]{
  \includegraphics[width=\textwidth,trim=0 0 30 20,clip]{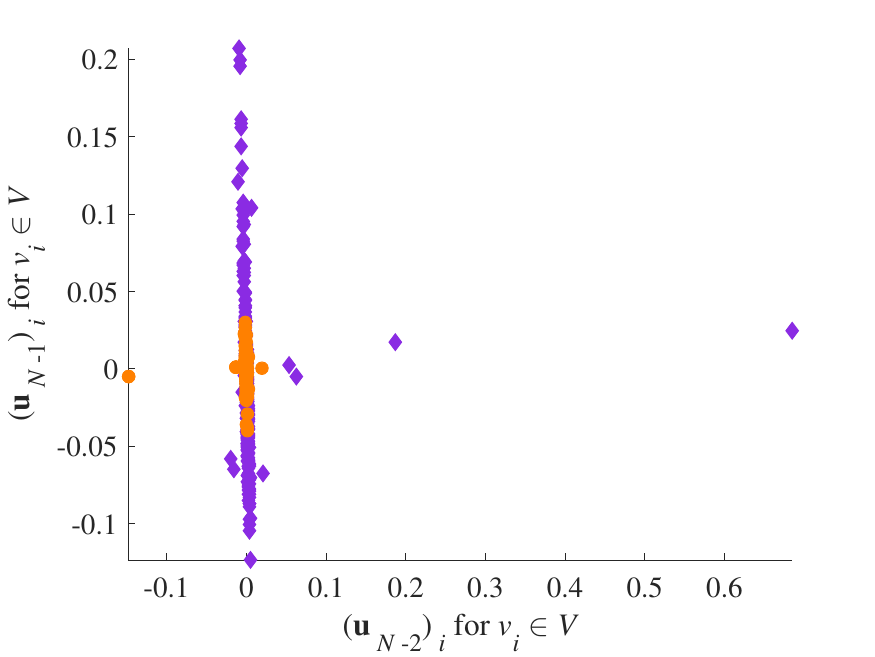}
  }
 \end{minipage}
 \begin{minipage}{0.19\linewidth}
 \subfigure[$\tau = 20$]{
  \includegraphics[width=\textwidth,trim=0 0 30 20,clip]{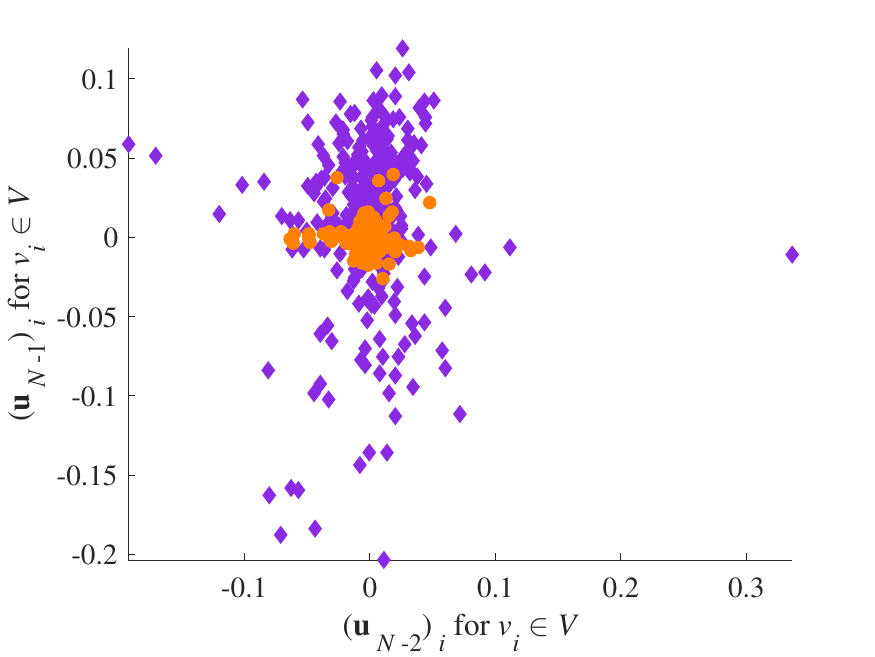}
  }
 \end{minipage}
 \begin{minipage}{0.19\linewidth}
 \subfigure[$\tau = 50$]{
  \includegraphics[width=\textwidth,trim=0 0 30 20,clip]{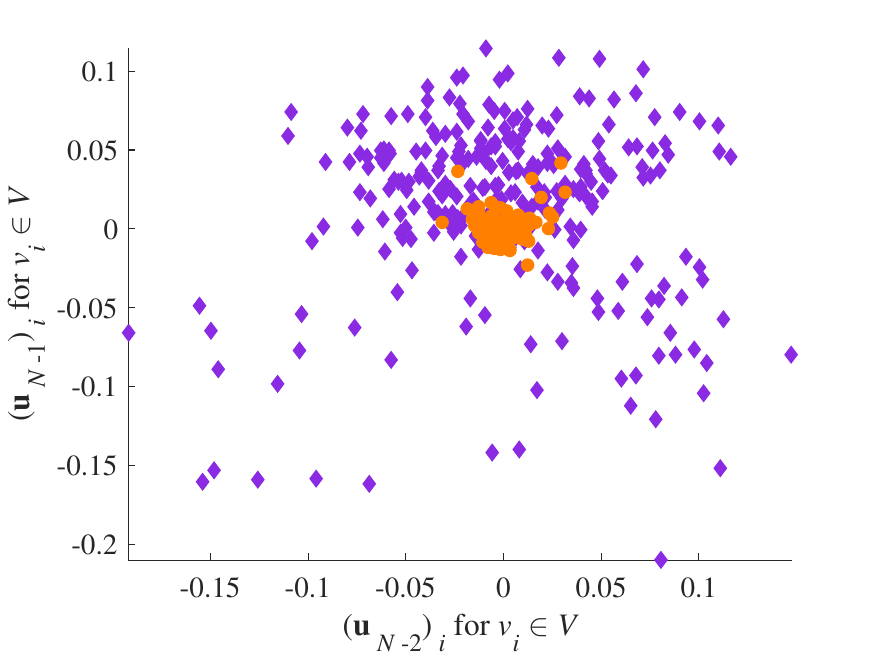}
  }
 \end{minipage}
 \begin{minipage}{0.19\linewidth}
 \subfigure[$\tau = 100$]{
  \includegraphics[width=\textwidth,trim=0 0 30 20,clip]{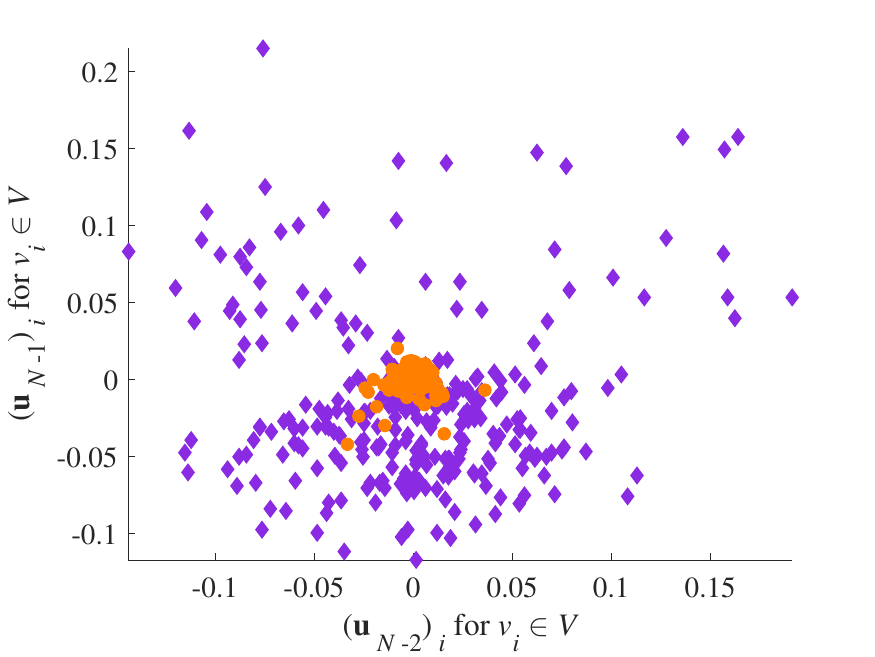}
  }
 \end{minipage}
 %----------
 \vspace{-0.2cm}
 \caption{Case study about high-frequency bases $[{\bf{\tilde u}}_{N-2}, {\bf{\tilde u}}_{N-1}]$ for two classes of USA w.r.t. $\tau \in \{ 0, 10, 20, 50, 100, 500\}$.
 }\label{Fig:Case-USA}
 \vspace{-0.2cm}
\end{figure*}

\section{Further Investigations about Node Identities and Positions}\label{App:Toy}

The full eigenvectors $[ {\bf{u}}_0, {\bf{u}}_1, \cdots, {\bf{u}}_{14}]$ w.r.t. normalized graph Laplacian ${\bf{L}}$ of the example graph in Fig.~\ref{Fig:Toy} is visualized in Fig.~\ref{Fig:Toy-1-vecs}.

\subsection{Investigation on Barbell Graphs}

In addition to Fig.~\ref{Fig:Toy}, we also validate our GSP-based investigation about node identities and positions on other graphs. The barbell graph \cite{ribeiro2017struc2vec} is a classic synthetic graph model to test the effectiveness of identity embeddings. In general, a barbell graph $B (n, c)$ consists of two complete graphs (with each having $n$ nodes) connected by a path with length $c$. Fig.~\ref{Fig:Toy-2} (a) gives an example of $B(6, 3)$ with $15$ nodes and corresponding ED on the normalized graph Laplacian ${\bf{L}}$, where each color denotes a unique structural role (i.e., identity). Fig.~\ref{Fig:Toy-2} (b) visualizes the full eigenvectors $[ {\bf{u}}_0, {\bf{u}}_1, \cdots, {\bf{u}}_{14} ]$.

Consistent with our observations in Fig.~\ref{Fig:Toy}, one can determine the community membership regarding node positions based on signs of entries in ${\bf{u}}_1$ w.r.t. the smallest non-zero eigenvalue $\lambda_1$ (i.e., the lowest non-trivial frequency). Concretely, we have $({\bf{u}}_1)_i < 0$ (or $>0$) for each node $v_i$ in community $1$ induced by $\{ v_1, \cdots, v_6 \}$ (or community $2$ induced by $\{ v_{10}, \cdots, v_{15} \}$).
In particular, $\{ v_7, v_8, v_9\}$ have values close to $0$ in ${\bf{u}}_1$, consistent with that it is unclear which community they belong to.

Furthermore, nodes with the same identity have exactly the same value in $\bf{u}_{14}$ w.r.t. the largest eigenvalue $\lambda_{14}$ (i.e., the highest frequency). For instance, red nodes $\{ v_1, \cdots, v_5, v_{11}, \cdots, v_{15}\}$ have the same value $-0.02$ and yellow nodes $\{ v_6, v_{10}\}$ have the same value $0.21$. This observation regarding node identities is consistent with the experiment results of some classic identity embedding methods (e.g., \textit{struc2vec} \cite{ribeiro2017struc2vec}).

In summary, Fig.~\ref{Fig:Toy-2} further validates that \textit{high- and low-frequency information in the graph spectral domain may capture node identities and positions, respectively}.

\subsection{Case Study on Real Graphs}

We further verify (\romannumeral1) the correlation between node identities and high-frequency information and (\romannumeral2) the effect of degree correction on real graphs.
For the \textit{USA} dataset with ground-truth of node identities, we set $\tau \in \{ 0, 10, 20, 50, 100\}$ and derived $[ {\bf{\tilde u}}_{N-2}, {\bf{\tilde u}}_{N-1} ]$ (i.e., bases w.r.t. the highest $2$ frequencies) for nodes of two sampled classes.
Fig.~\ref{Fig:Case-USA} visualizes the corresponding results, where each color denotes a unique class in \textit{USA}.

In Fig.~\ref{Fig:Case-USA} (a), most of the purple nodes tend to have larger values than orange nodes in ${\bf{u}}_{N-1}$ w.r.t. the highest frequency $\lambda_{N-1}$, which to some extent validate the correlation between high-frequency information and node identities.
However, there exist several outliers with extremely large or small values making this pattern less distinguishable.
In particular, with the increase of $\tau$, all the orange points in the visualized $2$D space tend to group together while purple points tend to move away from orange ones. We believe that such a pattern may help distinguish between the two classes. It also demonstrates that the newly introduced degree correction mechanism may further enhance the ability of RFA to capture node identities, which is consistent with our parameter analysis in Fig.~\ref{Fig:Param-tau}.

\section{Detailed Experiment Setup}\label{App:Exp-Set}

\textbf{Datasets}.
In the $10$ datasets (see Table~\ref{Tab:Data}), \textit{Europe}\footnote{https://github.com/leoribeiro/struc2vec/blob/master/graph/europe-airports.edgelist} and \textit{USA}\footnote{https://github.com/leoribeiro/struc2vec/blob/master/graph/europe-airports.edgelist} describe two aviation networks in Europe and the USA, with airports and routes between airports as nodes and edges.
\textit{Reality-Call}\footnote{https://github.com/cspjiao/RONE/blob/main/dataset/clf/reality-call.edge} was collected from phone call records of a small telephone network, where users and communication between users are treated as nodes and edges.
\textit{Actor}\footnote{https://github.com/cspjiao/RONE/blob/main/dataset/clf/actor.edge} and \textit{Film}\footnote{https://github.com/cspjiao/RONE/blob/main/dataset/clf/film.edge} record the co-occurrence relations between actors and keywords in a small Wikipedia dump.

\textit{PPI}\footnote{https://snap.stanford.edu/node2vec/Homo\_sapiens.mat} is a protein-protein interaction graph of Homo Sapiens.
\textit{BlogCatalog}\footnote{http://leitang.net/code/social-dimension/data/blogcatalog.mat} was extracted from social relationships provided by blogger authors.
\textit{Flickr}\footnote{http://leitang.net/code/social-dimension/data/flickr.mat} is a graph describing the contacts between users of a photo-sharing website.
\textit{Youtube}\footnote{https://snap.stanford.edu/data/com-Youtube.html} and \textit{Orkut}\footnote{http://leitang.net/code/social-dimension/data/blogcatalog.mat} were constructed based on the friendship relations in Youtube and Orkut.

For each dataset, we extracted the largest connected component to ensure that the graph topology is connected (i.e., with only one connected component) since we consider connected graphs as stated in Section~\ref{Sec:Prob}.

\textbf{Parameter Settings}.
The recommended parameter settings of \textbf{RFA(L)} and \textbf{RFA(H)} on corresponding datasets are depicted in Tables~\ref{Tab:Parm-L} and \ref{Tab:Parm-H}, where $d$ is the embedding dimensionality; $\tau$ is the degree correction term; $K$ is the number of iterations (i.e., GNN layers in RFA); $f_{\rm{act}} (\cdot)$ and `Norm' denote the settings of non-linear activation (i.e., $\tanh(\cdot)$ or $\exp (\cdot)$) and normalization (i.e., $l2$, z-score, or without normalization).

\end{document}